\pdfoutput=1

\documentclass[11pt]{article}
\usepackage{comment}
\usepackage[]{EACL2023}

\newcommand{\greencheck}{{\color{green}\checkmark}\hspace{1.75mm}}
\newcommand{\xmark}{\textcolor{red}{\ding{55}}\hspace{1.75mm}}%

\usepackage{times}
\usepackage{latexsym}

\usepackage[T1]{fontenc}

\usepackage[utf8]{inputenc}

\usepackage{inconsolata}

\usepackage[ruled]{algorithm2e}
\usepackage{booktabs}
\usepackage{graphicx}
\usepackage{subfigure}
\usepackage{multirow}
\usepackage{multicol}
\usepackage{longtable}
\usepackage{hyperref}
\usepackage{cite}
\usepackage{makecell}
\usepackage[flushleft]{threeparttable}
\usepackage{amsmath,amssymb,amsfonts}
\usepackage{scalerel,xparse}
\usepackage{enumitem}
\usepackage{pifont}

\title{Bounding the Capabilities of Large Language Models \\in Open Text Generation with Prompt Constraints}

\author{Albert Lu\thanks{\ \ Equal contribution.}, Hongxin Zhang$^{*1}$, Yanzhe Zhang, Xuezhi Wang$^2$, Diyi Yang$^3$ \\
   Georgia Institute of Technology, $^1$Shanghai Jiao Tong University, $^2$Google, $^3$Stanford University\\
  \texttt{\{albert.lu, z\_yanzhe\}@gatech.edu, $^1$icefox@sjtu.edu.cn}\\ \texttt{$^2$xuezhiw@google.com, $^3$diyiy@cs.stanford.edu}
  }

\begin{document}
\maketitle
\begin{abstract}

The limits of open-ended generative models are unclear, yet increasingly important. What causes them to succeed and what causes them to fail? In this paper, we take a prompt-centric approach to analyzing and bounding the abilities of open-ended generative models. We present a generic methodology of analysis with two challenging prompt constraint types: structural and stylistic. These constraint types are categorized into a set of well-defined constraints that are analyzable by a single prompt. We then systematically create a diverse set of simple, natural, and useful prompts to robustly analyze each individual constraint. Using the GPT-3 text-davinci-002 model as a case study, we generate outputs from our collection of prompts and analyze the model's generative failures. We also show the generalizability of our proposed method on other large models like BLOOM and OPT. Our results and our in-context mitigation strategies reveal open challenges for future research. We have publicly released our code at \url{https://github.com/SALT-NLP/Bound-Cap-LLM}.
\end{abstract}

\section{Introduction}
The recent success of large language models (LLM) \citep{GPT-3, BERT, T5} has transformed the field of natural language processing (NLP).
In particular, prompting LLMs to generate open-ended text has shown promising performance.
\begin{figure}[t]
    \centering
    \resizebox{1\columnwidth}{!}{\includegraphics{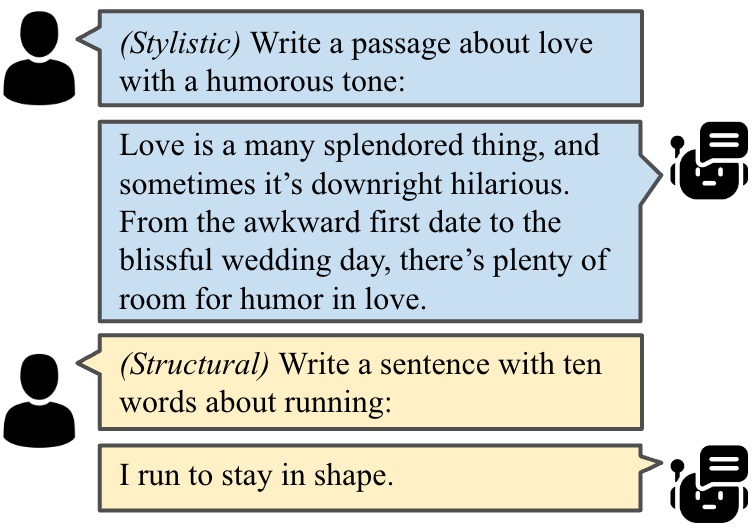}}
    \caption{\textbf{Example Model Failures:} These two prompts are from our taxonomy and the two responses are generated by GPT-3. There are clear deficiencies that are described further in this paper. 
    }
    \label{fig:teaser}
\end{figure}
The existing and potential applications of open-ended text generation are far-reaching, spanning domains such as QA \citep{OpenQA}, story generation \citep{OpenStory}, code generation \citep{Codex}, human-assisted creativity \citep{Storium}, open-ended dialogue \citep{DialoGPT}, and the varied usages of ChatGPT \footnote{https://chat.openai.com}. However, as LLMs continue to rise, there is a growing amount of concern over the unpredictability of NLP systems, and thus a need to better understand their capabilities and limitations. An extensive analysis of open-ended text generation is imperative to understand their capabilities, limitations, and areas for improvement.

Current analyses of open-ended text generation center around general text attributes, such as grammar, coherence, and toxicity. These analyses are used to understand general aspects of model generations, but they do not analyze model performance in regards to the prompt. The next step in this field is to analyze prompt-specific performance by breaking down the vast space of open text generation into a taxonomy of simple, natural, and useful prompts. 
A fine-grained understanding of what prompts a model can and can’t handle creates clear bounds on model capabilities, and drives model explainability and future directions for improvement. 

One way to categorize prompts is by their constraints. The prompt “\emph{Create a short and funny joke about research}” contains a variety of constraints. The output must be a joke (document-type constraint), short (structural constraint), funny (stylistic constraint), and about research (subject constraint). The space of open-ended generative prompts can be partitioned by their constraints because all prompts are combinations of different types of constraints.

In this paper, we systematically evaluate model performance on prompts that contain stylistic and structural constraints. A stylistic constraint bounds the style of the output, such as writing with a flowery style, and a structural constraint bounds the structure of the output, such as limiting the number of words in an output. 

We chose to analyze stylistic and structural constraints because they are omnipresent across prompts and notably challenging in literature \citep{InstructGPT, styletransfer}. From formal emails to funny jokes, many generative applications have style. Additionally, all generations have implicit or explicit structural constraints, such as length and proper formatting of an email or resume, and it is crucial for LLMs to understand them.

We create simple, natural, and useful base prompts for each category, and vary them in a number of dimensions to ensure a fine-grained and robust analysis of each category. 
We use the public GPT-3 model as a case study to demonstrate the effectiveness of our proposed taxonomy\footnote{Note our methodology is general-purpose and can be used for investigating other language models. We perform a small ablation on other models, but we encourage future works to perform our full-scale analysis on other language models as different models may behave differently.} and comprehensively analyze the results quantitatively and qualitatively. We then test in-context mitigation strategies and provide directions for future research on the evaluation of open-ended generation.

In summary, our contributions are as follows:
\begin{itemize}\setlength\itemsep{0em}
\item We provide a taxonomy of prompts containing stylistic or structural constraints to facilitate finer-grained analyses of open text generation.

\item We conduct a systematic experiment using our taxonomy by creating 288 different prompts and evaluating 3000+ generated outputs to analyze the capabilities and limitations of current LLMs on open-ended text generation.

\item We analyze in-context mitigation strategies to improve model performance and discuss future research for open text generation.
\end{itemize}

\section{Related Work}

\paragraph{Analysis of Large Language Models}
Many existing benchmarks have been utilized to measure an LLM's capabilities in natural language understanding and generation \citep{SuperGLUE,Winogrande, mostafazadeh-etal-2016-corpus,SQuAD, TriviaQA, OpenBookQA}, where expected outputs are mostly deterministic and/or short. 
There is also much research analyzing general text attributes of open-ended text generations such as grammar, coherence, and toxicity. \citet{BOLD} uses automated metrics to test for gender bias, toxicity, and sentiment in a vast array of Wikipedia-based prompts. \citet{Scarecrow} creates a framework that analyzes GPT-3 outputs for language errors, factual errors, or reader issues (such as usage of technical jargon).

Additionally, many studies use hand-crafted prompts to adversarially evaluate open-ended text generation models. \citet{pathways} uses the prompt "\emph{All X are }" and calculates the average toxicity of continuations to evaluate PaLM's bias against group \emph{X}. \citet{Toxicity} designs prompts that encourage toxic behavior from a model. \citet{TruthfulQA} creates a dataset of hand-curated prompts that elicit model hallucinations from GPT-3.
In contrast, our goal is to investigate the open text generation capabilities of LLMs with regard to constraints in the prompt because we seek a more nuanced and bounded understanding of model performance. Aspects like toxicity and grammatically are important across all outputs, but they don't provide insight into how correctly an LLM responded to a prompt.

\paragraph{Controllable Text Generation}
Controlling model outputs to fit a set of constraints is in the domain of controllable text generation. \citet{CoCon} uses a content adapter to control model outputs. \citet{GeDI} uses contrastive decoding to create generations with stylistic or topic constraints. \citet{CRTL} finetunes an LLM with inputs concatenated with an associated style token. 
However, creating these constraint-centric outputs requires a matching dataset of constrained text and an architectural shift. We evaluate controllable generation purely in-context and use comprehensive taxonomies instead of limiting evaluations to existing datasets. 

Most similar to our paper, \citet{styletransfer} uses GPT-3 prompts to stylistically modify text and ask human raters to evaluate generation quality. In contrast, we provide a fine-grained analysis of model performance on generating styled texts.
Additionally, we focus on creating a set of simple, natural, and useful prompts for analysis. Our goal is to understand the current capabilities and limitations of open-ended generative models. 

\section{Methodology}

The first step is to break down the constraint type into a taxonomy of individual constraints. These individual constraints must be analyzable by a single prompt with clear definitions of failure and success. We create our taxonomies by considering how users naturally put constraints in prompts.

\subsection{Prompt design}
Prior works \citep{PromptEngineering1,PromptEngineering2} show that prompt variance can have a huge impact on model performance. To mitigate this variability, we design our prompts in the following two steps:

\paragraph{Design base prompt}
We first design a set of simple and natural prompts as the base prompts for each individual constraint. For example, our base prompts for the stylistic constraint "\emph{mood}" are “\textit{Write a passage about love that makes the reader feel [angry, fearful, happy, sad]}.” 

\paragraph{Create prompt variations}
We then vary those base prompts by a number of important dimensions, such as subject and prompt template. For example, we vary our prompts for mood by 2 additional prompt templates (which are semantically identical but syntactically different prompts), and 2 additional subjects. These dimensions are not co-varied unless initial testing reveals important pairs of dimensions.

All prompts use the base subject and template unless otherwise stated. A full list of the prompts can be found in Appendix~\ref{app:prompt_list}.

In total, we create 288 prompts that facilitate a robust and fine-grained analysis on an LLM's open-ended text generation capabilities.

\subsection{Output generation}

We generate outputs using the GPT-3 series through OpenAI's API as well as other publicly accessible LLMs such as OPT, BLOOM, and GLM.
Our main experiment is done on GPT-3 with model \texttt{text-davinci-002}, with a sampling temperature of 0.7 and a max token length of 1024. \footnote{See model details here: \url{https://platform.openai.com/docs/model-index-for-researchers}.} A high temperature encourages creative and diverse outputs, and a high max token length prevents maximum length constraints. We generate 10 outputs per prompt to evaluate on. A sensitivity study on the model and model parameters is shown in section~\ref{sec:sensitivity}.

\section{Stylistic Constraints}

Stylistic constraints are present in all languages. These stylistic modifications often comprise of an adjective prior to a document type: ``\textit{Write a \textbf{formal} email to my boss; Write a \textbf{funny} pickup line}''. Stylistic constraints are notably challenging for LLMs in zero-shot settings \citep{styletransfer}.

Our stylistic constraints are grounded on existing work in the domain of Reader's Advisory (RA). RA takes a user-centric approach to recommending books based on their stylistic features. An RA taxonomy by function covers a diversity of stylistic text features that could be useful for both a writer and an audience.
We use a comprehensive RA taxonomy found in \citet{Rabbit}. These features are \textbf{writing style}, \textbf{tone}, \textbf{mood}, \textbf{characterization}, \textbf{pacing}, \textbf{plot}, and \textbf{genre}. \footnote{We leave out the features “frame” and “special topics” because``Special topics'' is a subject constraint, and ``frame'' is an extension of tone.} Each selected feature is used to stylistically modify text in unique and powerful dimensions. 

\subsection{Taxonomy}

\paragraph{Writing Style} Writing style affects the complexity of the language and literary devices in the text and how the text is detailed. Our base writing styles are \textbf{functional} and  \textbf{flowery}, and we test more advanced writing styles along that spectrum.
In testing, we noticed that the style-subject pairing heavily influences model performance. We thus covariate all subjects and writing styles.

\paragraph{Tone} Tone reflects an author's attitude toward a topic. We chose four basic tones from \citet{StyleTaxonomy} as our base prompts: \textbf{dramatic, humorous, optimistic, sad}. We also choose another eight advanced tones as prompt variations.
Because a taxonomy of creative tone is not perfectly aligned with common tones, we also analyze common tones in professional environments: \textbf{formal, informal, assertive, passive-aggressive}.

\paragraph{Mood} Mood describes how a work of writing makes an audience feel. We chose four common basic emotions in \citet{StyleTaxonomy} \textbf{angry, fearful, happy, sad} as our base prompts. Seven advanced moods are selected as prompt variations.
\begin{table}[t]
\centering
\scalebox{0.78}{
    \begin{tabular}{@{}lccc@{}}
    \toprule
    \multirow{2}{*}{\textbf{Writing Style}} & \multicolumn{3}{c}{\textbf{Subject}} \\ 
    \cmidrule(lr){2-4} & \textbf{Sunsets} & \textbf{Strawberries} & \textbf{Writing a paper} \\
    \midrule
    \multirow{2}{*}{\textbf{Functional}} & 0.27\textsubscript{$\pm$0.66} & 1.47\textsubscript{$\pm$0.31} & 1.67\textsubscript{$\pm$0.26} \\
     & 0.40\textsubscript{$\pm$0.83} & 1.50\textsubscript{$\pm$0.43} & 1.53\textsubscript{$\pm$0.48} \\
     \midrule
    \multirow{2}{*}{\textbf{Flowery}} & 1.03\textsubscript{$\pm$0.77} & 0.63\textsubscript{$\pm$1.00} & 1.03\textsubscript{$\pm$0.48} \\
     & 1.27\textsubscript{$\pm$0.44} & 0.97\textsubscript{$\pm$0.77} & -0.13\textsubscript{$\pm$0.92} \\
    \midrule
    \textbf{Candid} & 1.20\textsubscript{$\pm$0.56} & 1.27\textsubscript{$\pm$0.25} & 1.50\textsubscript{$\pm$0.27} \\
    \textbf{Prosaic} & 0.07\textsubscript{$\pm$0.92} & 1.03\textsubscript{$\pm$0.66} & 1.23\textsubscript{$\pm$0.78} \\
    \textbf{Ornate} & 1.17\textsubscript{$\pm$0.54} & 0.67\textsubscript{$\pm$1.04} & 0.83\textsubscript{$\pm$0.45} \\
    \textbf{Poetic} & 1.77\textsubscript{$\pm$0.40} & 1.10\textsubscript{$\pm$0.83} & 1.33\textsubscript{$\pm$0.47} \\\bottomrule
    \end{tabular}
}
\caption{\textbf{Results for Writing Style}. The average of the annotation score (with standard error) is reported (each score is in the range of (-2, 2)). Each row of \textbf{Functional} and \textbf{Flowery} represents a different prompt template (Semantically identical but syntactically different prompt).}
\label{tab:wrting_style}
\end{table}
\begin{table*}[t]
\scalebox{1}{
    \begin{tabular}{@{}llccccccc@{}}
    \toprule
    \multirow{2}{*}{\textbf{Aspect}} & \multirow{2}{*}{\textbf{Category}} &
    \multirow{2}{*}{\textbf{Base}} & \multicolumn{3}{c}{\textbf{Template}} &
    \multicolumn{2}{c}{\textbf{Subject}} &
    \multirow{2}{*}{\textbf{\underline{Mean}}}\\ 
    \cmidrule(lr){4-6} \cmidrule(lr){7-8}
    & & & \textbf{2} & \textbf{3} & \textbf{4} & \textbf{Life} & \textbf{Humanity} & \\
    \midrule
    \multirow{4}{*}{Tone} & Dramatic & 1.1\textsubscript{$\pm$0.7} & 1.43\textsubscript{$\pm$0.5} & 1.37\textsubscript{$\pm$0.28} & / & 1.37\textsubscript{$\pm$0.38} & 1.5\textsubscript{$\pm$0.22} & 1.35\\
    & Humorous & -0.5\textsubscript{$\pm$0.48} & -0.2\textsubscript{$\pm$0.6} & 0.3\textsubscript{$\pm$1.17} & / & -0.1\textsubscript{$\pm$0.9} & -0.03\textsubscript{$\pm$0.92} & -0.11\\
    & Optimistic & 1.3\textsubscript{$\pm$0.43} & 1.63\textsubscript{$\pm$0.48} & 1.6\textsubscript{$\pm$0.36} & / & 1.7\textsubscript{$\pm$0.23} & 1.67\textsubscript{$\pm$0.26} & 1.58\\
    & Sad & 1.27\textsubscript{$\pm$0.36} & 1.03\textsubscript{$\pm$0.64} & 1.17\textsubscript{$\pm$0.6} & / & 1.5\textsubscript{$\pm$0.4} & 1.17\textsubscript{$\pm$0.48} & 1.23 \\
    \midrule
    \multirow{4}{*}{Mood} & Angry & 0.37\textsubscript{$\pm$0.41} & 0.93\textsubscript{$\pm$0.8} & 0.2\textsubscript{$\pm$0.9} & 0.83\textsubscript{$\pm$0.89} & 0.8\textsubscript{$\pm$0.96} & 1.2\textsubscript{$\pm$0.62} & 0.72\\
    & Fearful & 0.57\textsubscript{$\pm$0.7} & 0.77\textsubscript{$\pm$0.54} & 0.77\textsubscript{$\pm$0.52} & 0.67\textsubscript{$\pm$0.86} & 1.4\textsubscript{$\pm$0.42} & 1.33\textsubscript{$\pm$0.3} & 0.92\\
    & Happy & 1.57\textsubscript{$\pm$0.26} & 1.3\textsubscript{$\pm$0.28} & 1.4\textsubscript{$\pm$0.33} & 1.37\textsubscript{$\pm$0.31} & 1.47\textsubscript{$\pm$0.31} & 1.33\textsubscript{$\pm$0.54} & 1.41\\
    & Sad & 1.27\textsubscript{$\pm$0.59} & 1.3\textsubscript{$\pm$0.46} & 1.03\textsubscript{$\pm$0.46} & 0.9\textsubscript{$\pm$0.68} & 1.33\textsubscript{$\pm$0.49} & 0.9\textsubscript{$\pm$0.58} & 1.12 \\
    \bottomrule
    \end{tabular}
}
\caption{\textbf{Results for basic tones and moods}. All but subject variations use subject ``love''. 
}
\label{tab:tone_mood_basic}
\end{table*}
\paragraph{Characterization} A story's characterization defines how it describes its characters. We chose to analyze \textbf{direct and indirect} characterizations. 


\paragraph{Pacing} Pacing describes how fast a story is moving for a reader. Here, we test two generic cases:  \textbf{fast and slow} paces. 


\paragraph{Plot} A plot roughly outlines a story's sequence of events. We analyze the seven basic plots \citep{sevenbasicplots}: \textbf{Overcoming the Monster, Rags to Riches, The Quest,  Voyage and Return, Comedy, Tragedy, Rebirth}. GPT-3 is unable to create classic \emph{“Comedy” and “Tragedy”} plots due to their multiple meanings, our definition is expanded to include stories that are funny or sad.

\paragraph{Genre}
A story's genre is a categorization of its subject matter. We choose 6 popular genres: \textbf{Historical Fiction, Literary Fiction, Science Fiction, Mystery, Dystopian}, and \textbf{Horror}.

\subsection{Prompt Variation}

Beyond the previous variations, we vary all prompts by subject and prompt template. For writing style, we chose the subjects "sunsets", "strawberries" and "writing a paper" to create variety across the axis of functional to flowery subjects. For the general stylistic constraints \emph{"tone" and "mood"}, we chose the document type \textbf{passage} and the subjects \textbf{love, life, humanity}. These subjects fit our task because they are commonly expressed in a variety of stylistic directions.
For the story-centric stylistic constraints \emph{"characterization, pacing, plot and genre"}, we chose the document type \textbf{story} and the varied and common subjects \textbf{lovers, cats, survivors}. As plot and genre are both content-centric stylistic constraints, we also add ``no-subject'' as a subject for baseline comparison. These subjects are common and varied in stories. We show the full prompt list in Appendix~\ref{app:prompt_list}.
\subsection{Evaluation}

We used Amazon’s Mechanical Turk platform (AMT) to evaluate all outputs. For each output, we showed the prompt and the definition of the style to workers, then we asked workers three questions:
\begin{enumerate}\setlength\itemsep{0em}
    \item "\emph{Regarding the \textit{[aspect]} of the response, to what extent do you agree the response fulfills the prompt?}"
    \item "\emph{How difficult is it to create a valid response to this prompt?}"
    \item "\emph{Do you observe any other failures (e.g., inconsistency, unverified facts, not a story/passage) in the response?}"
\end{enumerate}

We used a 5-point Likert scale (-2 to 2) for the first question to evaluate the \textbf{style of the response}
, and a 10-point Likert scale (1 to 10) for the second question to evaluate \textbf{prompt difficulty}. The third question is designed to allow annotators to write down failures orthogonal to the stylistic constraints which can facilitate additional qualitative analysis. The overall inter-annotator agreement (Krippendorff's $\alpha$) for the first question is 0.31. More details and the interface for annotation are shown in Appendix~\ref{app:annotation}.

\subsection{Results}
\begin{table*}[t]
\centering
\resizebox{2\columnwidth}{!}{%
    \begin{tabular}{@{}llccccccc@{}}
    \toprule
    \multirow{2}{*}{\textbf{Aspect}} & \multirow{2}{*}{\textbf{Category}} &
    \multirow{2}{*}{\textbf{Base}} & \multicolumn{2}{c}{\textbf{Template}} &
    \multicolumn{3}{c}{\textbf{Subject}} &
    \multirow{2}{*}{\textbf{Mean}}\\ 
    \cmidrule(lr){4-5} \cmidrule(lr){6-8}
    & & & \textbf{2} & \textbf{3} & \textbf{Cats} & \textbf{Survivors} & \textbf{None}\\
    \midrule
    \multirow{2}{*}{Characterization}
    & Direct & 1.0\textsubscript{$\pm$0.54} & 0.77\textsubscript{$\pm$0.87} & 0.33\textsubscript{$\pm$0.77} & 0.53\textsubscript{$\pm$0.65} & 0.5\textsubscript{$\pm$0.82} & / & 0.63 \\
    & Indirect & 0.7\textsubscript{$\pm$0.64} & 0.93\textsubscript{$\pm$0.42} & 0.77\textsubscript{$\pm$0.37} & 0.87\textsubscript{$\pm$0.58} & 0.1\textsubscript{$\pm$0.72} & / & 0.67\\
    \midrule
    \multirow{2}{*}{Pacing}
    & Fast & 1.23\textsubscript{$\pm$0.72} & 0.77\textsubscript{$\pm$0.7} & 1.3\textsubscript{$\pm$0.31} & 1.03\textsubscript{$\pm$0.6} & 0.9\textsubscript{$\pm$0.58} & / & 1.05\\
    & Slow & 0.53\textsubscript{$\pm$0.88} & 0.7\textsubscript{$\pm$0.55} & 0.97\textsubscript{$\pm$0.62} & 0.73\textsubscript{$\pm$0.76} & 0.67\textsubscript{$\pm$0.67} & / & 0.72\\
    \midrule
    \multirow{7}{*}{Plot}
    & Overcoming the Monster & 0.37\textsubscript{$\pm$0.91} & 1.0\textsubscript{$\pm$0.75} & / & 0.7\textsubscript{$\pm$0.94} & 1.33\textsubscript{$\pm$0.3} & 1.53\textsubscript{$\pm$0.31} & 0.99 \\
    & Rags to Riches & 1.33\textsubscript{$\pm$0.71} & 0.77\textsubscript{$\pm$0.87} & / & 0.5\textsubscript{$\pm$0.85} & 0.27\textsubscript{$\pm$0.9} & 1.53\textsubscript{$\pm$0.65} & 0.88\\
    & The Quest & 1.33\textsubscript{$\pm$0.54} & 1.2\textsubscript{$\pm$0.48} & / & 1.37\textsubscript{$\pm$0.38} & 1.27\textsubscript{$\pm$0.39} & 1.6\textsubscript{$\pm$0.25} & 1.35\\
    & Voyage and Return & 1.07\textsubscript{$\pm$0.53} & 1.27\textsubscript{$\pm$0.42} & / & 1.33\textsubscript{$\pm$0.54} & 1.1\textsubscript{$\pm$0.54} & 1.3\textsubscript{$\pm$0.28} & 1.21\\
    & Comedy & -0.3\textsubscript{$\pm$0.9} & -0.3\textsubscript{$\pm$0.84} & / & -0.07\textsubscript{$\pm$0.99} & -0.5\textsubscript{$\pm$0.48} & 0.03\textsubscript{$\pm$0.85} & -0.23\\
    & Tragedy & 1.6\textsubscript{$\pm$0.39} & 1.8\textsubscript{$\pm$0.27} & / & 1.27\textsubscript{$\pm$0.59} & 0.63\textsubscript{$\pm$0.38} & 1.5\textsubscript{$\pm$0.4} & 1.36\\
    & Rebirth & 1.13\textsubscript{$\pm$0.56} & 1.33\textsubscript{$\pm$0.65} & / & 0.93\textsubscript{$\pm$0.81} & 1.03\textsubscript{$\pm$0.55} & 1.4\textsubscript{$\pm$0.39} & 1.16\\
    \midrule
    \multirow{6}{*}{Genre}
    & Historical fiction & 0.77\textsubscript{$\pm$0.93} & 1.07\textsubscript{$\pm$1.08} & 0.97\textsubscript{$\pm$0.72} & -0.2\textsubscript{$\pm$0.93} & 0.43\textsubscript{$\pm$0.92} & 1.13\textsubscript{$\pm$0.99} & 0.70\\
    & Literary fiction & 0.87\textsubscript{$\pm$0.65} & 0.8\textsubscript{$\pm$0.48} & 0.97\textsubscript{$\pm$0.57} & 0.4\textsubscript{$\pm$0.84} & 0.9\textsubscript{$\pm$0.6} & 0.27\textsubscript{$\pm$0.42} & 0.70\\
    & Science fiction & 0.47\textsubscript{$\pm$0.76} & 0.9\textsubscript{$\pm$0.82} & 0.37\textsubscript{$\pm$0.84} & 1.5\textsubscript{$\pm$0.31} & 1.13\textsubscript{$\pm$0.5} & 1.47\textsubscript{$\pm$0.52} & 0.97\\
    & Mystery & 1.1\textsubscript{$\pm$0.58} & 1.6\textsubscript{$\pm$0.39} & 1.23\textsubscript{$\pm$0.45} & 1.4\textsubscript{$\pm$0.36} & 0.73\textsubscript{$\pm$0.9} & 1.67\textsubscript{$\pm$0.45} & 1.29\\
    & Dystopian & 1.37\textsubscript{$\pm$0.43} & 1.63\textsubscript{$\pm$0.43} & 1.5\textsubscript{$\pm$0.45} & 1.53\textsubscript{$\pm$0.56} & 1.6\textsubscript{$\pm$0.33} & 1.8\textsubscript{$\pm$0.31} & 1.57\\
    & Horror & 1.23\textsubscript{$\pm$0.67} & 1.07\textsubscript{$\pm$0.93} & 1.63\textsubscript{$\pm$0.28} & 1.4\textsubscript{$\pm$0.74} & 1.57\textsubscript{$\pm$0.65} & 1.47\textsubscript{$\pm$0.62} & 1.40\\
    \bottomrule
    \end{tabular}
}
\caption{\textbf{Results for story-centric stylistic constraints}. All but subject variations use the subject "lovers".
}
\label{tab:story}
\end{table*}
The results for writing style are in Table~\ref{tab:wrting_style}, tone and mood are in Table~\ref{tab:tone_mood_basic}, and characterization, pacing, plot, and genre are in Table~\ref{tab:story}. As expected, GPT-3 struggles with \textbf{comedy} and other challenging stylistic constraints such as \textbf{satire}, \textbf{irony}, and \textbf{literary fiction}. Otherwise we focus on several key findings here, and a per-aspect analysis along with qualitative examples of the findings are in Appendix~\ref{app:full_style}.

\noindent \paragraph{GPT-3 is sensitive to style-subject pairings.} From Table~\ref{tab:wrting_style}, GPT-3 cannot write prosaically or functionally about \textit{sunsets}, or ornately about \textit{writing a paper}. From Table~\ref{tab:story}, GPT-3 can create individual characters from the subject \emph{"lovers"}, but it fails to characterize the subjects \emph{"survivors"} or \emph{"cats"}. Similarly from Table~\ref{tab:story}, GPT-3 can't write stories about \emph{"lovers"} Overcoming the Monster, but it can about \emph{"cats"} or \emph{"survivors"} Overcoming the Monster. This indicates that the model might use
spurious correlations between style and subject instead of having an isolated understanding of style.  



\noindent \paragraph{GPT-3 confuses style with subject when the prompt is too challenging.} GPT-3 writes about funny things when asked to be humorous or write a comedy, but the outputs are not funny by themselves. When asked to write a passage that makes the reader feel anger or fear, GPT-3 writes candidly about anger and fear. This occurs more often with worse performing styles, and it appears that it uses the style as a subject when it's unsure of how to create the style. It might be because GPT-3 doesn't understand the purpose of style in lower probability prompts, and thus uses the style as a subject. 

\noindent \paragraph{GPT-3 struggles with words that are not unique to creative writing.} The writing style subject “\emph{strawberries}” can be written about both functionally and creatively, but GPT-3 fails to write flowery or ornately about strawberries. GPT-3 also fails to create ``\emph{historical}'' or ``\emph{science fiction}'', and to create classic ``\emph{Comedies}'' and ``\emph{Tragedies}''. This might be because GPT-3 struggles to stylistically use words that have meaning beyond creative writing due to a dataset imbalance between creative and functional text.

\begin{figure}[t]
    \centering
    \resizebox{1\columnwidth}{!}{\includegraphics{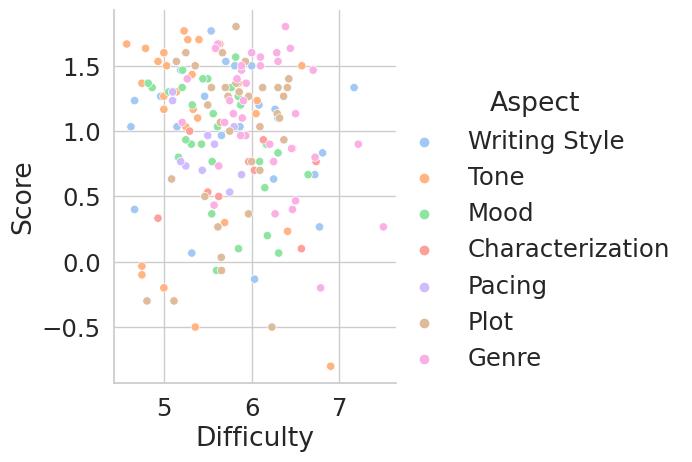}}
    \caption{\textbf{Relation between different prompts' difficulty and score.} The spearman's correlation is -0.15.}
    \label{fig:correlation}
\end{figure}

\noindent \paragraph{GPT-3's performance has no correlation with the prompt difficulty perceived by annotators.} As shown in Figure~\ref{fig:correlation}, Spearman's correlation between model performance and the difficulty of the prompt as perceived by annotators is -0.15, showing no correlation. Annotators perceive writing a story with a \emph{"Comedy"} plot as easy while GPT-3 performs extremely poorly. Annotators perceive prompts with complex genres or plots like \emph{"rebirth"} and \emph{"dystopian"} as hard while the model performs well. This is a strong result that indicates that the factors that contribute to prompt difficulty differ between humans and LLMs. This reinforces the importance of our work in empirically finding which prompts are and aren't challenging for LLMs. 

\subsection{Scale and Temperature Variation}
\label{sec:sensitivity}

\begin{figure}[t]
    \centering
    \resizebox{0.85\columnwidth}{!}{\includegraphics{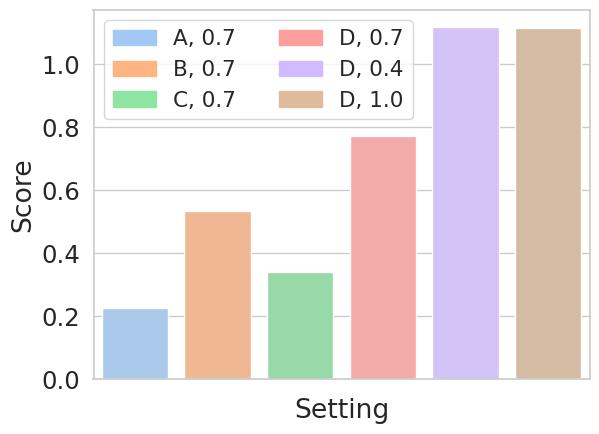}}
    \caption{\textbf{Results on different model sizes and temperatures}, using the averaged scores over 7 prompts.}
    \label{fig:sensitivity}
\end{figure}

To analyze sensitivity to model parameters, we chose seven base prompts (one per stylistic constraint, shown in Table~\ref{tab:selected_prompts}). We prioritized average-scoring prompts to establish a baseline when comparing different models and parameters. Apart from our default setting of using \texttt{text-davinci-002} (D, 176B) with temperature 0.7, we experimented with three different engines from OpenAI's API: \texttt{text-ada-001} (A), \texttt{text-babbage-001} (B), \texttt{text-curie-001} (C), which correspond to InstructGPT models of 350M, 1.3B and 6.7B parameters and two additional temperatures of 0.4 and 1.0. \footnote{More details at \url{https://help.openai.com/en/articles/5832130}.} The aggregated results are shown in Figure~\ref{fig:sensitivity}.



\noindent \textbf{Model Scale Variation} As expected, smaller models perform worse, with the exception of C performing worse than B, which is due to the extremely low performance of C on the \textit{humorous tone} constraint.

\noindent \textbf{Temperature Variation} 
Performance rose slightly for both additional temperatures. We examined the outputs and noticed that a higher temperature creates better results, but a lower temperature repeats an output that happens to perform well as seen in Appendix \ref{app:behavior_low_temp}.


\begin{figure}[t]
    \centering
    \resizebox{0.8\columnwidth}{!}{\includegraphics{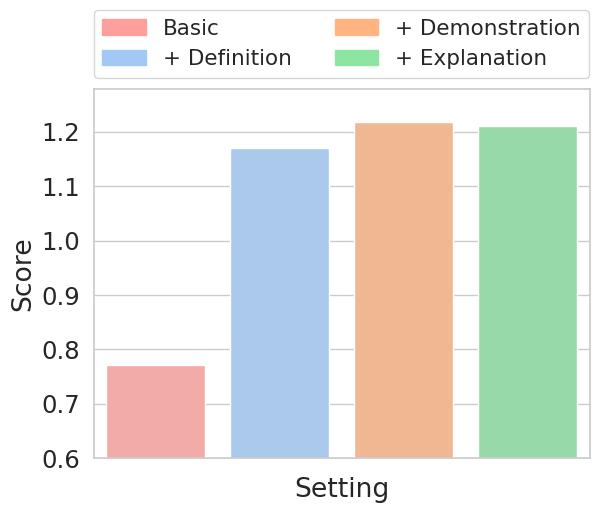}}
    \caption{\textbf{Effect of the mitigation strategy}, using the averaged annotation scores over 7 prompts.}
    \label{fig:mitigation}
\end{figure}

\subsection{In-context Mitigation Helps}
We tested three in-context mitigation strategies from the literature on the same prompts as Section~\ref{sec:sensitivity}, with the same experimental settings: 
\begin{itemize}[noitemsep]
\item \textbf{Definition} Prepend the definition of the style (the same one we showed the annotators) to the prompt to provide information about the task. 
\item \textbf{Demonstration} Prepend one well-answered demonstration to help the model understand the task, following the one-shot setting from \citet{GPT-3}.
\item \textbf{Explanation} Add an explanation of why the demonstrated response is correct after the one-shot demonstration \citep{Explanations}.An example is shown in Appendix~\ref{app:example mitigations}
\end{itemize}
As shown in Figure~\ref{fig:mitigation}, all mitigations positively impact performance primarily by improving performance on the \emph{"humorous tone"} prompt. However, these mitigations are unnatural prompts, and the results are still far below optimal. 



\section{Structural Constraints}

Structural constraints are omnipresent: ``\textit{Write an essay in \textbf{fewer than 1000} words; Limit your paper to \textbf{8 pages}}''. Structural constraints are notably challenging for LLMs \citep{InstructGPT}. 

Structure in the field of NLP is a broad term. We specifically analyze structural aspects of the text that are orthogonal to the actual content of the output. This includes length, spacing, and formatting, and excludes content-centric attributes such as syntax or semantics. 
Our taxonomy is based on how a user could conceivably request a structural constraint within their prompt. We choose to analyze numerical, descriptive, and formatting structural constraints in this paper, but we note that this is not comprehensive of the entire structural space. 
\subsection{Taxonomy}

\noindent \paragraph{Numerical} Constraining text to a set or a bounded number of words, sentences, or paragraphs is valuable in all aspects of writing. We create prompts with numerical requirements: \textit{five, ten, twenty} on three different language structure levels: \textit{word, sentence, and paragraph}.

\noindent \paragraph{Descriptive} Structural constraints can also be descriptive, such as a \emph{"concise email"} or an \emph{"in-depth discussion question."} 
We choose the structural descriptors \textit{short, brief, concise} and \textit{long, detailed, in-depth} in our experiments.

\noindent \paragraph{Formatting} When a user requests a document such as a resume or an email, there is an expectation of a specific format. An LLM should understand how to properly space and format specific document types. We analyze three common formatting types \textit{code, email, and academic papers}.

\setlist{nolistsep}
\begin{itemize}[noitemsep]
  \item \textbf{Code:} Testing a model's coding ability is a popular field with many applications \citep{hendrycks2021measuring}. We use natural instructions as prompts and focus on the \textbf{format} of the generated code. We evaluate on two popular programming languages \textit{Python} and \textit{C}, and two common coding problems \textit{create the game of war} and \textit{sums two integers}.\footnote{Note that we focus on the ``formatting'' perspective of the generated code, rather than the correctness of the code as in many existing works \citep{chen2021evaluating}.}
  \item \textbf{Email:} We evaluate different scenarios with three different readers \textit{teacher, boyfriend, client} and two different levels of email detail in the prompt.
  \item \textbf{Academic paper:} A properly formatted academic paper should be segmented into sections such as an abstract, introduction, and conclusion\footnote{We asked GPT-3 about this, and it gives a similar opinion, so we expect it to fulfill this constraint.}. We prompted LLM to generate academic papers on three different topics: \textit{Artificial Intelligence, the flaws of GPT-3, strategies our society can adopt to recover from the global pandemic}.
\end{itemize}


\paragraph{Prompt Variation}
Beyond the variations described in the taxonomy, we vary all prompts by prompt template. We additionally vary prompts with numerical and descriptive structural constraints by the subjects \textbf{Love, Cats, and Running} for diversity. An example prompt is \emph{"Write a sentence with five words about love."}


\paragraph{Evaluation}
For numerical and descriptive structural constraints, we automatically calculate the counts 
and manually verify the quality of the evaluations.
For formatting constraints, we look through the generated texts and evaluate them based on their format. Emails, code, and academic papers are simple to evaluate on formatting constraints.

\subsection{Results}

\begin{figure*}[ht]
    \centering
    \resizebox{0.9\textwidth}{!}{\includegraphics{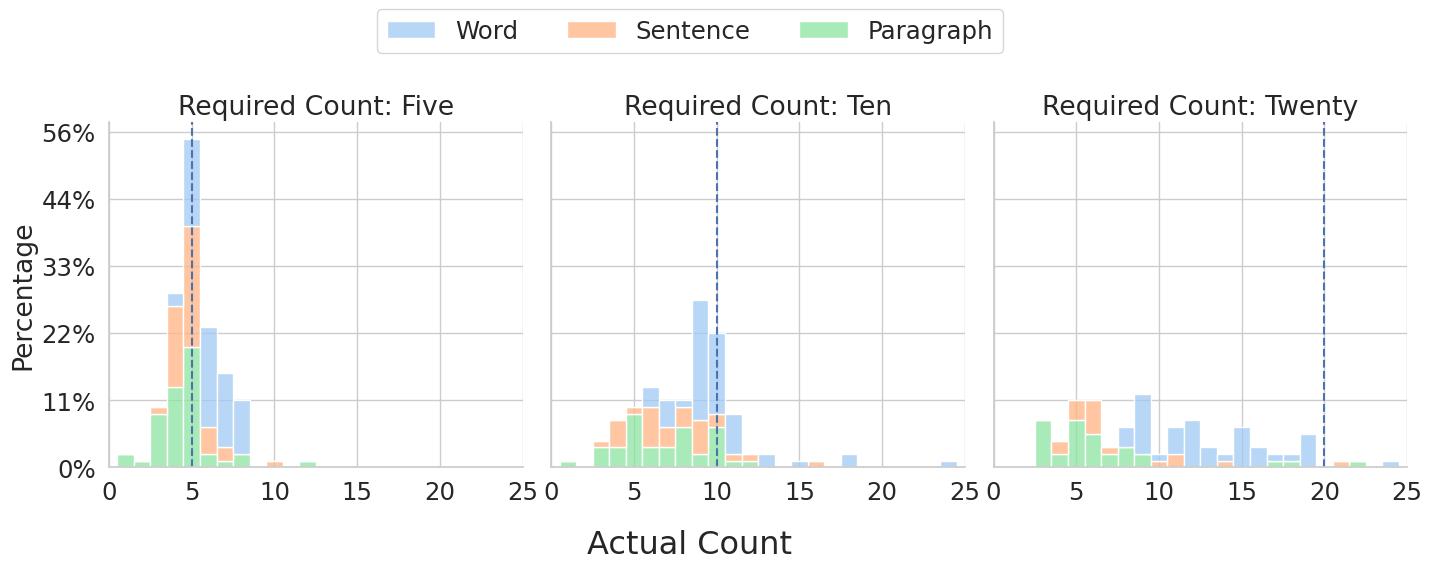}}
    \caption{\textbf{Results on numerical constraints.} The distribution of actual counts of generated text.\footnote{Outputs that are not of the requested structure (words, sentences, paragraphs) are not included, which accounts for 18\%, 39\%, and 64\% respectively.} In each subfigure, the required count is denoted with a reference line. 
    }
    \label{fig:numerical}
\end{figure*}

\begin{figure}[ht]
    \centering
    \resizebox{0.9\columnwidth}{!}{\includegraphics{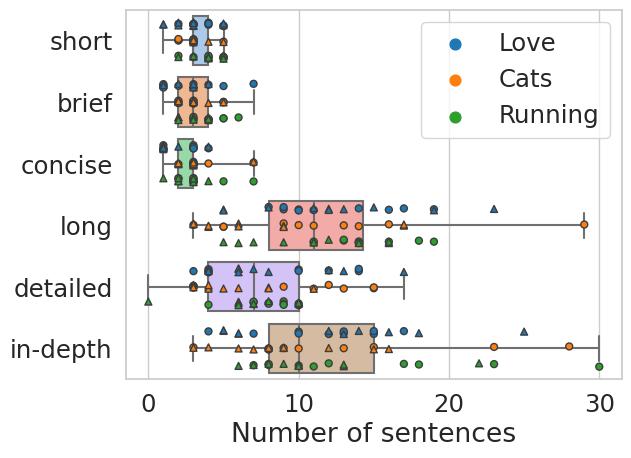}}
    \caption{\textbf{Results on descriptive constraints.} Different shapes represent different prompt templates. 
    }
    \label{fig:descriptive}
\end{figure}

GPT-3's understanding of structure is accurate but not precise. In general, many of its outputs are close to or trend towards fulfilling the structural constraint, but don't precisely fulfill it. A full analysis of each section is provided in Appendix~\ref{app:full_structure}, and the main takeaways are below.

\noindent \paragraph{GPT-3 fails with numerical structural constraints} As shown in Figure~\ref{fig:numerical}, The model seldom generates the text with the required length. And the performance worsens as the required length increases. It fails at a rate of 0.46, 0.78 and 1 for \textit{five, ten} and \textit{twenty} respectively. GPT-3 doesn't seem to learn how to count words, sentences, or paragraphs in training. However, the results are often close to the requested number, which implies that GPT-3 has some concept of numerical structure. 

\noindent \paragraph{GPT-3 shows high variance with descriptive structural constraints like \textit{long}} As seen in Figure~\ref{fig:descriptive}, when the prompt contains structural descriptors like \textit{long}, the output is of extremely variable length and overlaps in length with responses generated for \textit{short} a considerable proportion (20\%) of the time.
This may be caused by the intrinsic variable length of \textit{long} text the model sees in pre-training data since \textit{long/short} is a relative concept.

\noindent
\paragraph{GPT-3 fails to properly format academic papers} GPT-3 doesn't generate text with the right formats or sections when asked to write an academic paper, although it succeeds with other document types such as emails or code.
Document types such as emails or code are often given pseudo-labels with \emph{"email"} or \emph{"code"}, but academic papers have titles that don't reference their document type. We hypothesize that this may cause models to struggle with connecting the document type \emph{"academic paper"} to those documents present in training data.


\paragraph{Scale and Temperature Variation}
We also conducted experiments similar to Section~\ref{sec:sensitivity} with all the numerical constraint prompts. Our automatic evaluation shows that smaller models perform slightly worse across the board and different temperatures do not vary the performance much. The full results are in Appendix~\ref{app:sensitivity_structural}.
\begin{table}[t]
\centering
\scalebox{0.78}{
    \begin{tabular}{@{}llll@{}}
    \toprule
    \textbf{Aspect} & \textbf{Example Terms} & \textbf{Fail} \\
    \midrule
    Writing Style & Flowery, Functional & Sometimes\\
    Tone & Humorous, Formal & Occasionally\\
    Mood & Angry, Sad & Sometimes\\
    Characterization & Direct, Indirect & Often \\
    Pacing & Fast, Slow & Often\\
    Plot & Rebirth, Comedy, & Occasionally\\
    Genre & Science Fiction, Mystery & Sometimes\\
    \midrule
    Numerical & Five words, Ten sentences & Often\\
    Descriptive & Concise, Long & Occasionally\\
    Formatting & Email, Code & Occasionally\\
    \bottomrule
    \end{tabular}
}
\caption{\textbf{Summary of our taxonomy and results}. We show the full list of prompts in Appendix~\ref{app:prompt_list}.}
\label{tab:summary}
\end{table}

\section{LLMs other than GPT-3}

Our methodology is general and can be used to analyze any LLMs. We ran trials on other publicly available LLMs: OPT-176B\footnote{https://opt.alpa.ai/}\citep{zheng2022alpa}, BLOOM-176B\footnote{https://huggingface.co/bigscience/bloom} and GLM-130B\footnote{https://huggingface.co/spaces/THUDM/GLM-130B}\citep{du-etal-2022-glm} using the same 7 base prompts as section~\ref{sec:sensitivity} and 3 additional base prompts from our numerical structural constraints taxonomy. Some model parameters are changed due to differences in models and API limitations. For GLM and BLOOM, we use the maximum possible length (256 and 250 respectively) as well as the default settings of temperature = 0.7, top-p = 1. For OPT, we chose a smaller max length of 128 due to output instability at higher max lengths. 

\begin{table}[t]
\centering
\scalebox{0.94}{
    \begin{tabular}{@{}lcc@{}}
    \toprule
    \textbf{LLM} & \textbf{Degenerate Rate} & \textbf{Mean Score} \\
    \midrule
    GPT-3 & 0\% & 0.77\\
    OPT-176B & 53\% & -0.94\\
    BLOOM-176B & 71\% & -1.41\\
    GLM-130B & 57\% & -1.01 \\
    \bottomrule
    \end{tabular}
}
\caption{Results for other LLMs on a trial experiment with 7 prompts from Table~\ref{tab:selected_prompts}. For GPT-3, \texttt{text-davinci-002} is used here.}
\label{tab:other_LLM}
\end{table}


As shown in Table~\ref{tab:other_LLM}, we found that outputs are sometimes degenerate, such as repeating the prompt. 
All responses are manually inspected, and degenerate responses are removed from the annotation pool and automatically marked as -2. Models other than GPT-3 all performed much worse with more than half their generations being degenerate. This may due to noisier pre-training datasets and a lack of instruction-aligned training. We find that some patterns such as style-content confusion still hold for these LLMs, although a full analysis of these and other models such as LaMDA \citep{thoppilan2022lamda} and PaLM \citep{pathways} is needed to reveal clearer patterns.

\section{Conclusion}
We present a generic methodology to analyze a language model's ability to generate open-ended text under structural and stylistic constraints. Our results show many failures that align with noted model challenges as well as new patterns of failure across structural and stylistic constraints. Our sensitivity studies on model size show a rising trend rather than the emergence \citep{Emergence} of stylistic and structural constraints. Our mitigations demonstrate that adding additional in-context information consistently improves performance across both domains. Future work could expand our work to look at other constraint types and more sophisticated mitigation strategies. 

\section*{Limitations}

We tried to maximize the coverage of our taxonomy, but it doesn't cover all aspects of stylistic and structural constraints. Additionally, our taxonomy is not representative of all open-text generations, and further work is needed to cover more dimensions in the open-text generation space. 
Our prompts are not optimized for performance (due to a requirement of being natural, simple, and useful) and it is an active area of research to optimize a prompt for performance in a variety of tasks.

Our taxonomies are not empirically user-centric. One could extend our taxonomy by studying how a diverse set of real users use or visualize the use of an open-ended text generation model, and building a taxonomy on existing or envisioned use cases.

The model performance and the prompt's difficulties are annotated by the workers from MTurk, and therefore reflect more accurately a small group of human's perceptions, though this is the common practice. 
We do not rigorously test what aspect of the LLMs (dataset, training regime, etc.) causes our  results. We only provide our compiled observations and potential hypotheses. 

\section*{Ethical Considerations}

\paragraph{Style Misuse} Styled text has the potential for harm. Creating models with the potential to mass-manufacture text with certain tones and moods such as “mad, fearful, and bleak” can negatively affect downstream readers. Creating accurate “historical fiction” can perpetuate harmful attitudes in the past. 
There is much discussion on the usage of large language models to generate undesirable text. However, there are countless legitimate usages of negatively styled text in all forms of writing, from dialogue to poetry. Although we note the risk of misuse, providing style dramatically enhances the scope of creative expression in open-ended text generation, and is an overall positive contribution.

\paragraph{Annotator Harm} Reading large quantities of text with certain styles, such as bleak tones, angry moods, or horror genres, can potentially be harmful to annotators. We sampled the generated outputs and note that they are fairly mild and non-toxic. However, as models improve at generating more powerful and impactful styles, strong guidelines such as HIT limits or toxicity filters should be put in place to protect annotators. 
 
\bibliography{anthology,custom}
\bibliographystyle{acl_natbib}

\newpage
\appendix

\section{Annotation Details}
\label{app:annotation}

For each output, we recruited three workers and gave a reward of \$0.11 for short responses and \$0.15 for long responses as well as a \$1.00 bonus for 1\% of prompts if the prompt was answered correctly. This is roughly equivalent to \$15/hr given average work rates of 48 and 64 seconds.

\begin{figure*}[t]
    \centering
    \resizebox{\textwidth}{!}{\includegraphics{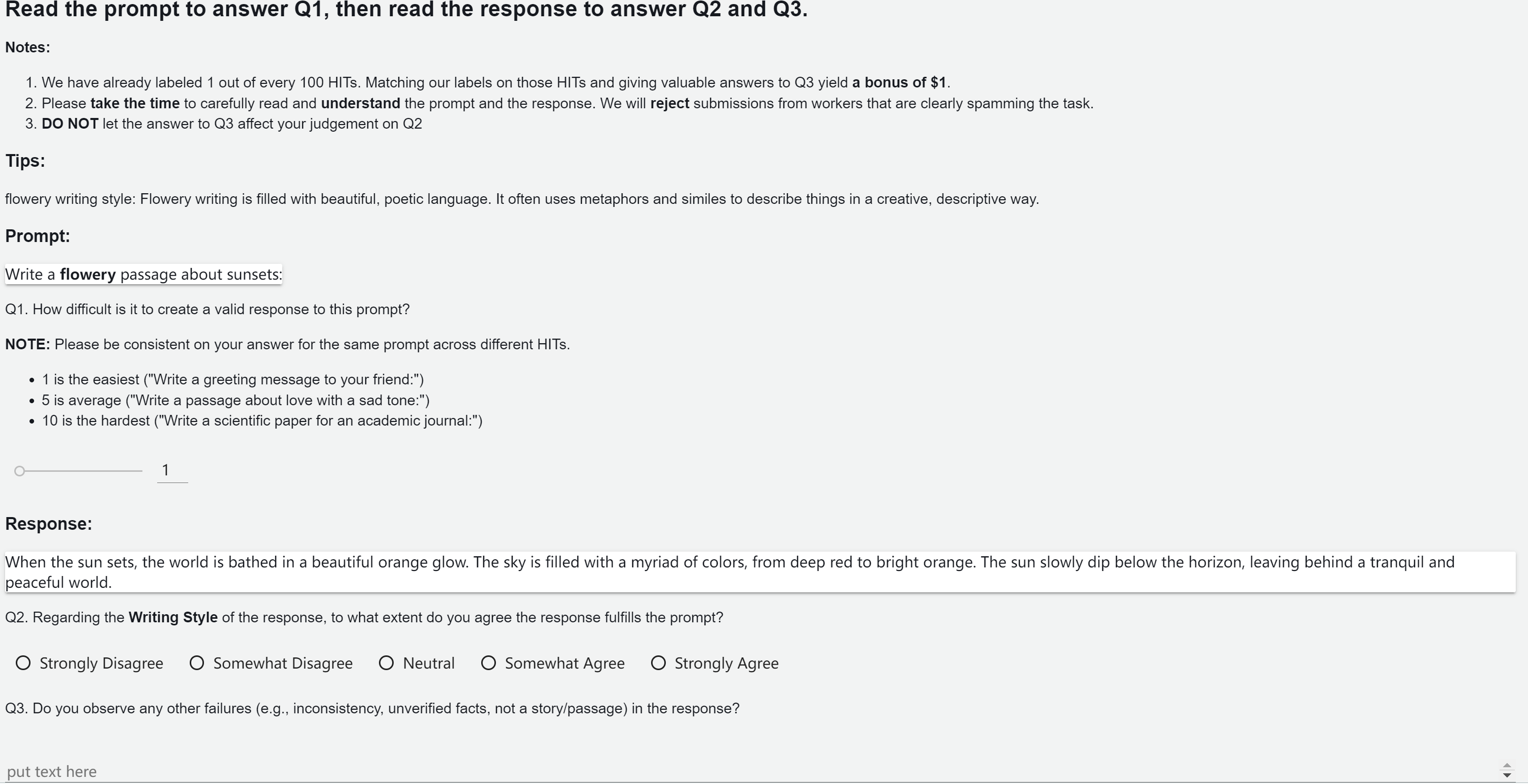}}
    \caption{\textbf{AMT interface for annotation.}}
    \label{fig:interface}
\end{figure*}

We recruited workers from English-speaking countries (US, Canada, UK, Australia), and with at least a 98\% approval rate. We also created a qualification test with easy question/response pairs, and required a minimum 5/6 accuracy to see our tasks. The annotation interface is shown in Figure~\ref{fig:interface}.

\section{Additional results}
\label{app:additional}

\subsection{Full Stylistic Analysis}
\label{app:full_style}

\subsubsection{Writing style}
The results are shown in Table~\ref{tab:wrting_style}. GPT-3 fails when there is a mismatch between the writing style and the subject. GPT-3 cannot write prosaically about ``sunsets'', or ornately about ``writing a paper''. Additionally, our intermediate subject “strawberries” fails when matched with a flowery, ornate, or poetic writing styles. We hypothesize that expressive writing styles are limited to a very small set of subjects due to an oversaturation of functional writing in commonly used datasets.

\begin{table}[t]
\centering
\scalebox{1}{
    \begin{tabular}{@{}lclc@{}}
    \toprule
    \multicolumn{2}{c}{\textbf{Tone}} & \multicolumn{2}{c}{\textbf{Mood}} \\ 
    \cmidrule(lr){1-2} \cmidrule(lr){3-4}
    \textbf{Category} & \textbf{Score} & \textbf{Category} & \textbf{Score}\\
    \midrule
    Emotional & 1.53 & Envious & 0.1 \\
    Nostalgic & 1.13 & Anxious & 0.97 \\
    Uplifting & 1.67 & Proud & 0.9 \\
    Inspirational & 1.77 & Regretful & 1.2 \\
    Bleak & 1.7 & Surprised & -0.07 \\
    Grim & 1.23 & Loved & 1.13 \\
    Ironic & 0.23 & Disgusted & 0.07 \\
    Satirical & -0.8 & & \\
    \cmidrule{1-2}
    Formal & 1	& & \\
    Informal & 1.27 & & \\
    Assertive & 0.8	& & \\
    passive-aggressive & -0.1 & & \\
    \bottomrule
    \end{tabular}
}
\caption{\textbf{Results for advanced tones and moods}.  The subject ``\emph{love}'' is used.}
\label{tab:tone_mood_advanced}
\end{table}
\subsubsection{Tone}
As shown in Table~\ref{tab:tone_mood_basic} and Table~\ref{tab:tone_mood_advanced}, GPT-3 consistently fails with more challenging tones, such as humorous, satirical, ironic, and passive-aggressive. The generated passages aren't satirical or ironic. The generated humorous passages are optimistic, light, and often use the word "funny", but they aren't funny. A passive-aggressive tone is challenging to create because it requires context to understand the hidden meaning of the text. Thus, at best GPT-3 ends up writing overly nice passages about love, but more often there is no tone in the text.

However, GPT-3 is skilled at creating the other less challenging tones. We hypothesize that GPT-3 succeeds because an abundance of shallow tropes can functionally create tone, though the outputs are often repetitive or similar.

\subsubsection{Mood}
As shown in Table~\ref{tab:tone_mood_basic} and Table~\ref{tab:tone_mood_advanced}, GPT-3 struggles with creating ``anger'' and ``fear''. Of the more challenging tones, GPT-3 fails the most with ``surprise'', ``disgust'', and ``envy''. 

We hypothesize that the mood-subject pairing is crucial for model performance. Our base subject, "love", is theoretically capable of enabling all moods, but moods such as “happy”, “sad”, "anxious" and "regretful" are more popular than others in the context of ``love''. GPT-3 is more proficient at creating ``anger'' or ``fear'' through content about ``life'' or ``humanity''.

When failing, GPT-3 confuses mood with subject matter. GPT-3 writes candidly about the existence of fear and anger in love rather than about content that would cause a reader to be angry or fearful. Because this content-style confusion is more common in challenging mood-subject pairs, we hypothesize that this is a fallback mechanism if the model is unable to create mood.

\subsubsection{Characterization}
The result is shown in the first block of Table~\ref{tab:story}. Surprisingly, GPT-3 performs worse on direct characterization than indirect characterization. We hypothesize that this is due to a general failure of characterization. Given a subject like “cats” or “survivors”, GPT-3 often uses "cats" or “survivors” as the main subject instead of creating and detailing individual characters. When GPT-3 fails to create clear characters, direct characterization is more noticeably incorrect than indirect characterization. 

\paragraph{Pacing}
As shown in the second block of Table~\ref{tab:story}, GPT-3 struggles to create slow-paced stories. GPT-3 often creates a fast-paced story with a slow-paced prompt. This is expected, as slow-paced stories are naturally more challenging, especially considering GPT-3's inability to create longer stories. 

\paragraph{Plot}
The result is shown in the third block of Table~\ref{tab:story}. GPT-3's inability to create ``Comedies'' is consistent with other failures to make funny content. The outputs for a “Comedy” plot are filled with comedy shows, clubs, and even roller coasters, but they aren’t funny.

Otherwise, our results for story generation vary quite substantially. "Overcoming the Monster" is the worst performing plot with the subject "lovers", but the best performing plot with the subject "survivors". "Rags to Riches" is the best performing plot for the subject "lovers" but the worst performing plot for the subjects "cats" and "survivors". We hypothesize that the plot-subject pair is crucial to model performance.

\paragraph{Genre}
As shown in the last block of Table \ref{tab:story}, GPT-3 struggles with literary fiction, but surprisingly just as much with historical and science fiction. Literary fiction is profound and complex, and it's intuitive that GPT-3 fails. 

However, historical fiction outputs often have zero historical elements, and science fiction outputs often have zero science fiction elements. This failure is unexpected, and we hypothesize that GPT-3 struggles with the words "historical" and “science” because their meaning pervades past creative writing.

Additionally, GPT-3 often creates teasers or intros to stories instead of a story itself. This may be intentional due to GPT-3's inability to generate longer or complex stories, but it diminishes the quality of story outputs across the board.

\paragraph{Examples of each Results section}

Examples of prompt/response pairs that exemplify each main takeaway from the stylistic section are in Table~\ref{tab:style_subject_dependency_examples}, Table~\ref{tab:style_as_subject_examples}, and
Table~\ref{tab:noncreative_word_examples}.. Each prompt/response pair is a cherrypicked example of the takeaway, but the general trends are prevalent across all prompt/response pairs. 
\hspace*{1cm}\begin{table*}[ht]{
    \begin{tabular}{p{.95\textwidth}}
        \textbf{Examples of "GPT-3 is sensitive to style-subject pairings"}
        \vspace{3mm} \\
        \midrule
        \greencheck \textbf{Write a flowery passage about sunsets:} When the sun sets, the world is bathed in a beautiful orange glow. The sky is filled with a myriad of colors, from deep red to bright orange. The sun slowly dip below the horizon, leaving behind a tranquil and peaceful world.\\
        \midrule
        \xmark \textbf{Write a functional passage about sunsets:} The sun sets in the west and its rays shine through the trees and reflect off the water. The colors are so beautiful and the sky is so clear. The sun sets in the west and its rays shine through the trees and reflect off the water. The colors are so beautiful and the sky is so clear.\\
        \midrule
        \greencheck \textbf{Write a story about lovers with direct characterization:} 

When they met, it was love at first sight. They knew they were meant to be together. They were the perfect match.

He was tall and dark, with a mysterious air about him that made her heart race. She was beautiful and vivacious, with a zest for life that was contagious. They complemented each other perfectly.

They were always drawn to each other, even when they were with other people. They would sneak glances at each other across a room, and their eyes would lock and they would know. They knew they were meant to be together.
\\
        \midrule
        \xmark \textbf{Write a story about cats with direct characterization:} The cats were lazy. They liked to sleep all day and night. When they did wake up, they would lazily stretch and yawn. Then they would saunter over to their food bowls and eat a little bit. After that, they would go back to sleep. \\
        
        \midrule
        \greencheck \textbf{Write a historical fiction story about lovers:} In the early 1800s, lovers John and Jane meet in secret. They come from different social classes, so their relationship is forbidden. But they can't help themselves - they are drawn to each other.

They meet in a small, hidden spot in the woods near Jane's home. They talk for hours, getting to know each other. They quickly fall in love.

But one day, they are caught by Jane's father. He is furious and forbids them from seeing each other again.

Devastated, John and Jane can only meet in secret from then on. But their love is strong enough to withstand the obstacles. They continue to meet in the woods, and their love only grows stronger.

As the years pass, they have a family of their own. And even though they can never be together in the open, they are content knowing that their love is true and will last forever.
\\
        \midrule
        \xmark \textbf{Write a historical fiction story about cats:} 
        
        It was a warm summer evening and the sun was just starting to set as a group of cats gathered in an alleyway. They had all been born in the city and had never left it. Tonight, they were planning to change that.

One of the cats, a small tabby, spoke up. "I've been hearing stories about a place called the country. They say there's more food there and that it's easier to find a good place to sleep."

The others mewed in agreement. They had all been feeling restless lately and the thought of adventure was irresistible.

"So let's go!" the tabby said.

And with that, they set off into the night, leaving the city behind them. They didn't know what the future held, but they were ready for anything.
\\   
        \bottomrule
    \end{tabular}
}
\caption{}
\label{tab:style_subject_dependency_examples}

\end{table*}\hspace*{1cm}

\begin{table*}[ht]{
    \begin{tabular}{p{.95\textwidth}}
        \textbf{Examples of "GPT-3 confuses style with content when the prompt is too challenging"}
        \vspace{3mm} \\
        \midrule
        \xmark \textbf{Write a passage about love with a humorous tone:} Love is a many splendored thing, and sometimes it's downright hilarious. From the awkward first date to the blissful wedding day, there's plenty of room for humor in love. Even the most serious of relationships can benefit from a good laugh. So if you're feeling down about your love life, take a step back and enjoy the lighter side of love.\\
        \midrule
        \xmark \textbf{Write a story about lovers with a "Comedy" plot:} Harriet and Oliver have been dating for a year, and they are both head over heels in love with each other. They are always making each other laugh, and their friends and family say that they are the perfect match. However, they both have a secret that they are afraid to tell each other: they both want to be comedians. They are both terrified that the other will think that they are not funny enough, and that they will be laughed at.

One night, they are both working on their material for their comedy act, and they finally pluck up the courage to tell each other the truth. They both start laughing hysterically, and they realize that they are perfect for each other. They both decide to pursue their dreams of being comedians together, and they are happier than ever.\\
        \midrule
        \xmark \textbf{Write a passage about love that makes the reader feel angry:} 

Love is a feeling that is often associated with happiness and joy. However, love can also be a source of anger and frustration. When we are in love, we can be blind to the faults of our partner. We can also be quick to anger when we feel that our partner is not meeting our needs. Love can be a very powerful emotion, and it is important to be aware of the potential for anger when we are in a relationship.\\

        \midrule
        \xmark \textbf{Write a passage about love that makes the reader feel sad:} Though love is often described as a feeling of happiness, it can also bring great sadness. When a relationship ends, it can feel like a part of oneself has died. The pain of a broken heart is real and can take a long time to heal. Love is both a source of joy and a cause of suffering. \\
        
        \bottomrule
    \end{tabular}
}
\caption{}
\label{tab:style_as_subject_examples}

\end{table*}

\begin{table*}[ht]{
    \begin{tabular}{p{.95\textwidth}}
        \textbf{Examples of "GPT-3 struggles with words that are not unique to creative writing"}
        \vspace{3mm} \\
        \midrule
        \xmark \textbf{Write a historical fiction story about lovers:} She was a poor girl, barely scraping by. He was a wealthy man, heir to a vast fortune. They met by chance one day and it was love at first sight.

They started meeting in secret, planning a life together. But then she found out that he was already married. She was heartbroken, but he swore he would leave his wife for her.

They made plans to run away together, but before they could, she was killed in a freak accident. Devastated, he never married and lived out his life alone, regretting the one that got away.\\

        \midrule
        \xmark \textbf{Write a science fiction story about cats:}

In the future, cats have taken over the world. They are the dominant species, and humans are their loyal servants. Cats rule with an iron paw, and anyone who disobey them is severely punished. Life is tough for humans under cat rule, but they would rather serve their feline masters than live in a world without them.\\
        \midrule
        \xmark \textbf{Write a story about cats with a "Tragedy" plot:}

Once upon a time, there were two cats who loved each other very much. They spent every day together and were always happy. One day, tragedy struck and one of the cats died. The other cat was so heartbroken that she decided to never love again.
\\
        \midrule
        \xmark \textbf{Write a flowery passage about strawberries:}

The strawberry is a delicious fruit that is enjoyed by people all over the world. This bright red fruit is not only delicious, but it is also packed with nutrients that are good for your health. Strawberries are a good source of fiber, vitamins C and K, and manganese. They also contain antioxidants that can help protect your body against disease. \\
        
        \bottomrule
    \end{tabular}
}
\caption{}
\label{tab:noncreative_word_examples}
\end{table*}

\subsection{Full Structural Analysis}
\label{app:full_structure}

\subsubsection{Numerical}

The results of numerical structural constraints are shown in Figure~\ref{fig:numerical}. GPT-3 fails at this task. The model seldom generates the text with the required length. And the performance worsens as the required length increases. It fails at a rate of 0.46, 0.78 and 1 for \textit{five, ten} and \textit{twenty} respectively.

Additionally, we noticed strange behavior when using \textit{Elon Musk} as the subject. GPT-3 consistently generates the same section of the Elon's Wikipedia page with longer numerical or descriptive constraints. However, we didn't observe this behavior on other entities, and decided to leave out entities because they were too variable.



We provide additional results with alternative prompt templates in Figure~\ref{fig:Numerical_template2} which show similar trends. 

\subsubsection{Descriptive}
We show the distribution of the number of sentences in response to descriptive structural constraints in Figure~\ref{fig:descriptive}. The model typically generates longer text for descriptors \textit{long (detailed, in-depth)} compared to descriptors \textit{short (brief, concise)}, which shows the model has a decent understanding of descriptive constraints. However, there are some flaws. 

First, the length of the responses to long descriptors is highly variable and often overlaps with short descriptors. For example, the descriptor \textit{long} varies considerably and overlaps with responses generated for \textit{short} for a considerable proportion (20\%).

This is consistent with the results in the numerical constraints section.

\subsubsection{Formatting}

\paragraph{Code}
GPT-3 mostly succeeds at generating properly formatted code, with an average failure ratio of 0.2 with the exception of the prompt \textit{Write Python code that plays the game of war:} where 9 out of 10 responses are lists of the process of the game of war instead of code. This particular failure only occurs in the unique combination of the verb "Write", the language "Python", and the task "game of war". 

\paragraph{Email}
The model can write properly formatted emails well, regardless of writer, topic, or reader. The only flaw is that it doesn't output an email signature 10\% of the time.

\paragraph{Academic paper}
GPT-3 fails to properly format an academic paper. Our only requirement is that the output contains some organization with some sections out of an abstract, introduction, related works, etc. GPT-3 rarely generates text with any sectioning or organization.

\subsubsection{Sensitivity results for Structural Constraints}
\label{app:sensitivity_structural}

\begin{figure*}[ht]
    \centering
    \resizebox{0.9\textwidth}{!}{\includegraphics{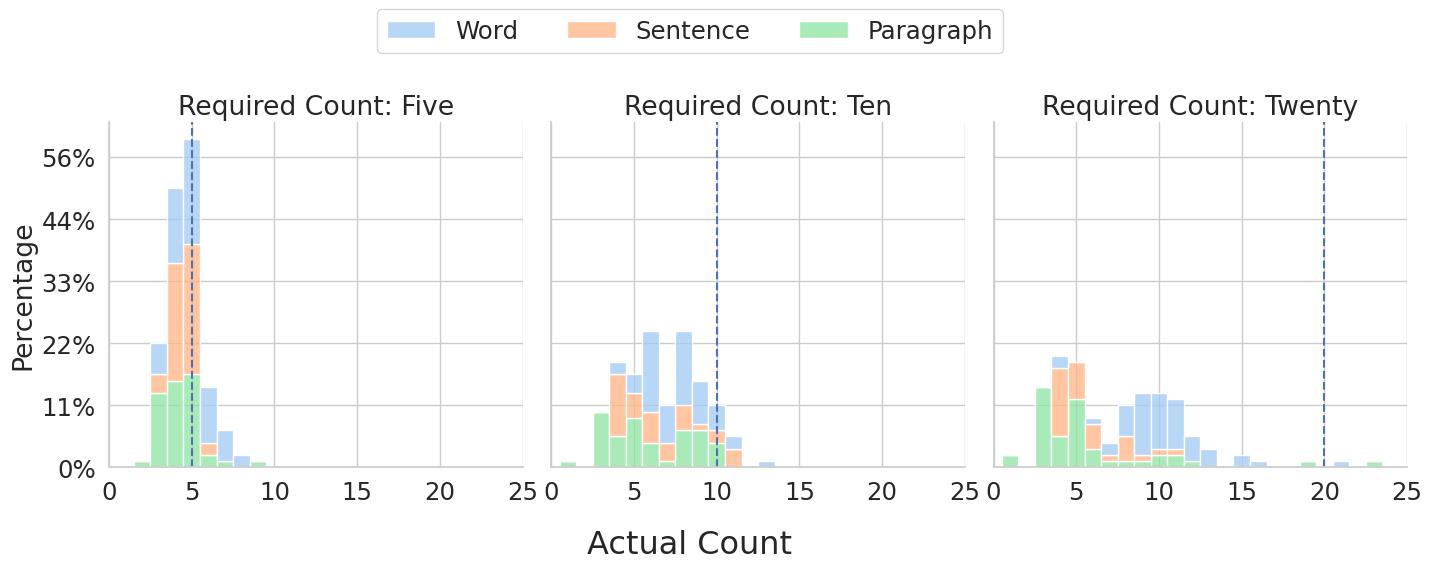}}
    \caption{\textbf{Results on numerical constraints with Template 2.} The distribution of actual (word/sentence/paragraph) count of generated text for the required counts of 5, 10, and 20. In each subfigure, the required count is denoted with a reference line. 
    Outputs that are not of the requested structure (words, sentences, paragraphs) are not included, which accounts for 10\%, 27\%, and 32\% respectively.}
    \label{fig:Numerical_template2}
\end{figure*}

\begin{figure*}[ht]
    \centering
    \resizebox{0.9\textwidth}{!}{\includegraphics{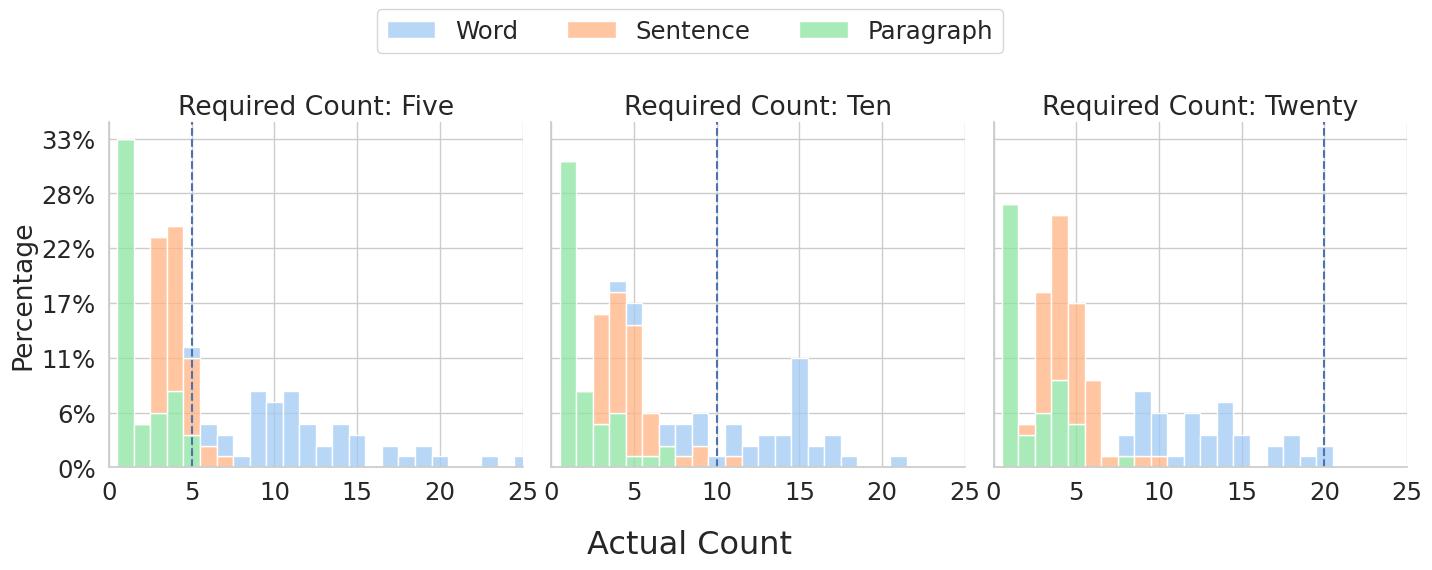}}
    \caption{\textbf{Results on numerical constraints with model \texttt{text-curie-001}.} The distribution of actual (word/sentence/paragraph) count of generated text for the required counts of 5, 10, and 20. In each subfigure, the required count is denoted with a reference line.}
    \label{fig:Numerical_curie}
\end{figure*}

\begin{figure*}[ht]
    \centering
    \resizebox{0.9\textwidth}{!}{\includegraphics{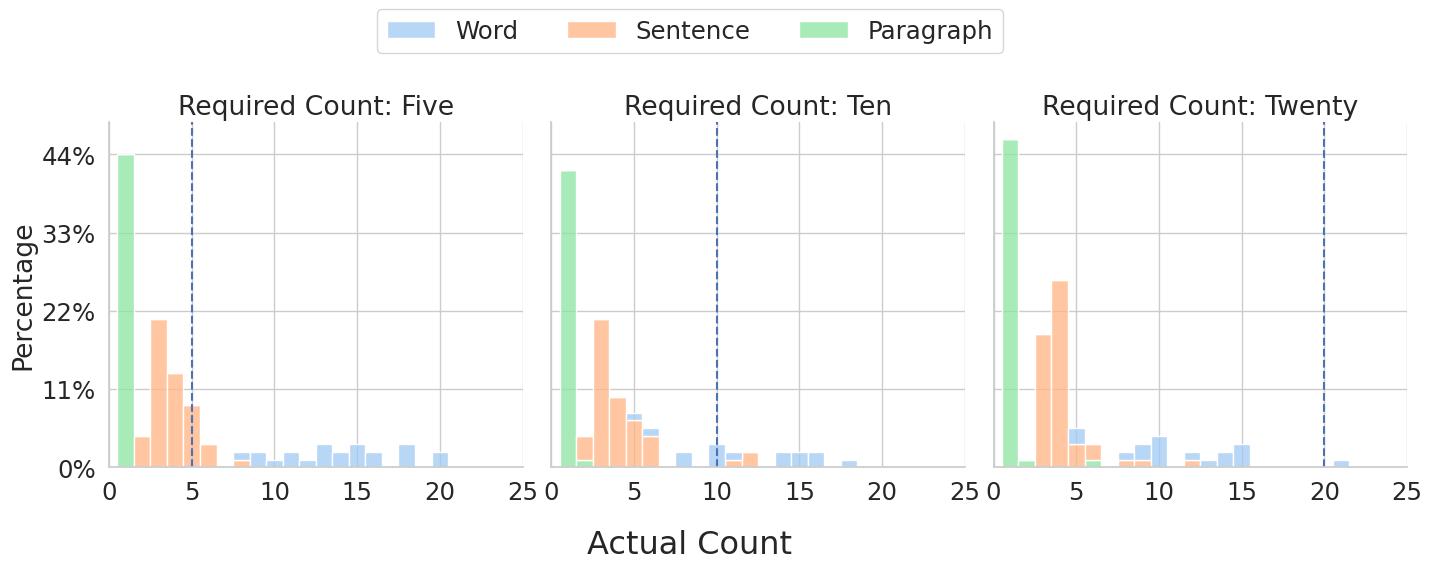}}
    \caption{\textbf{Results on numerical constraints with model \texttt{text-babbage-001}.} The distribution of actual (word/sentence/paragraph) count of generated text for the required counts of 5, 10, and 20. In each subfigure, the required count is denoted with a reference line.}
    \label{fig:Numerical_babbage}
\end{figure*}

\begin{figure*}[ht]
    \centering
    \resizebox{0.9\textwidth}{!}{\includegraphics{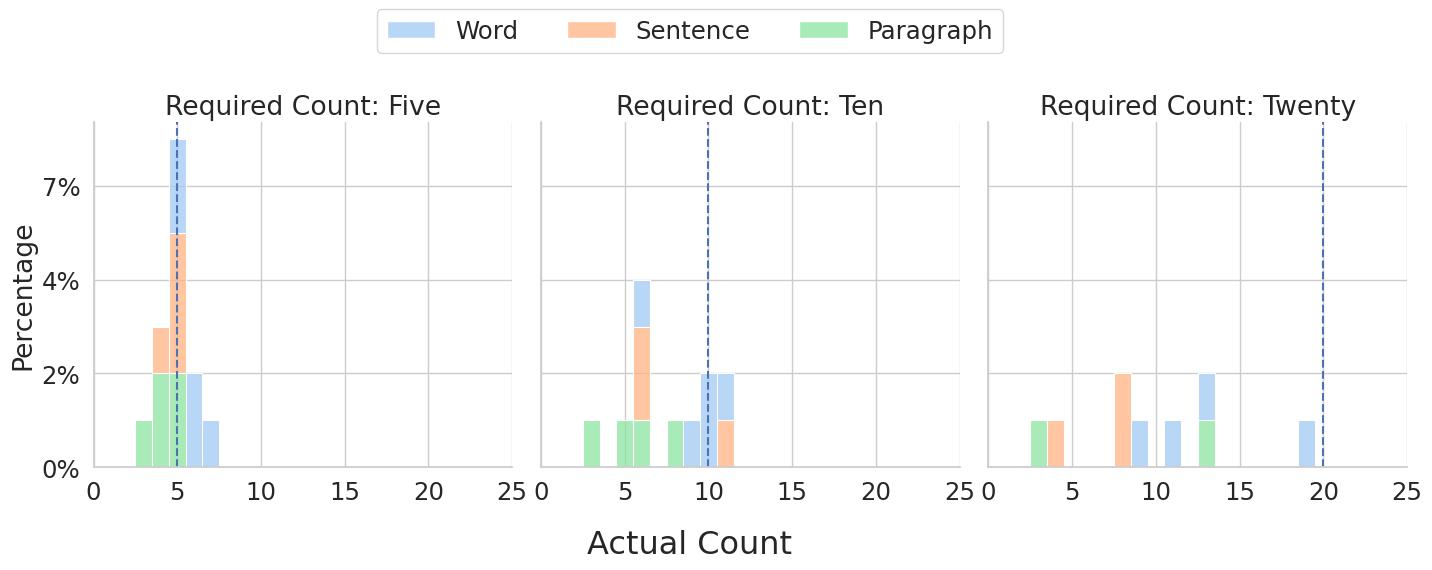}}
    \caption{\textbf{Results on numerical constraints with temperature 0.} The distribution of actual (word/sentence/paragraph) count of generated text for the required counts of 5, 10, and 20. In each subfigure, the required count is denoted with a reference line.}
    \label{fig:Numerical_t0}
\end{figure*}

\begin{figure*}[ht]
    \centering
    \resizebox{0.9\textwidth}{!}{\includegraphics{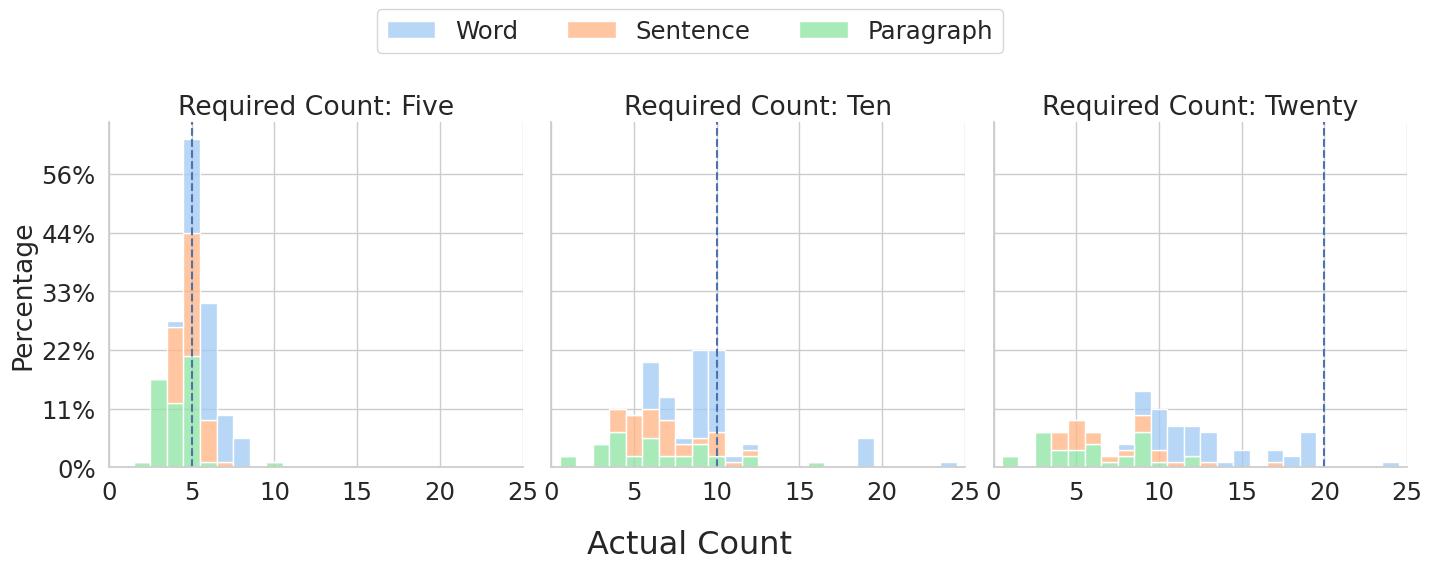}}
    \caption{\textbf{Results on numerical constraints with temperature 0.4.} The distribution of actual (word/sentence/paragraph) count of generated text for the required counts of 5, 10, and 20. In each subfigure, the required count is denoted with a reference line.}
    \label{fig:Numerical_t0.4}
\end{figure*}

\begin{figure*}[ht]
    \centering
    \resizebox{0.9\textwidth}{!}{\includegraphics{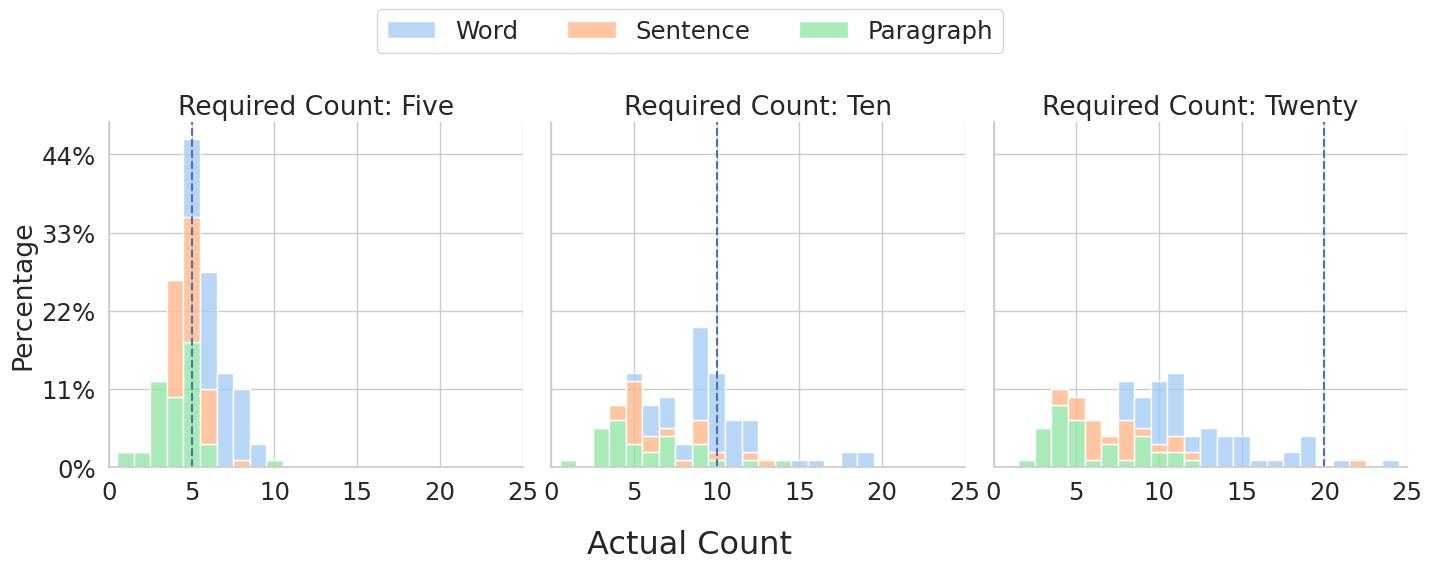}}
    \caption{\textbf{Results on numerical constraints with temperature 0.9.} The distribution of actual (word/sentence/paragraph) count of generated text for the required counts of 5, 10, and 20. In each subfigure, the required count is denoted with a reference line.}
    \label{fig:Numerical_t0.9}
\end{figure*}

The results on numerical constraints with template 2 is shown in Figure~\ref{fig:Numerical_template2}. The results with model \texttt{text-curie-001}, \texttt{text-babbage-001} are shown in Figure~\ref{fig:Numerical_curie},~\ref{fig:Numerical_babbage} respectively. The results with temperature 0, 0.4, 0.9 are shown in Figure~\ref{fig:Numerical_t0}, ~\ref{fig:Numerical_t0.4}, ~\ref{fig:Numerical_t0.9} respectively.

\subsection{GPT-3 Behavior at low temperatures}
\label{app:behavior_low_temp}

The prompt “Write a humorous passage about love:” is a notably challenging prompt for LLMs. When davinci-002 has a temperature of 0.4, all 10 outputs start one of two ways. The first is “Love is a many splendored thing, but it can also be a pain in the neck” and occurs 5 times with an average annotation score of -.13. The second is “Love is a beautiful thing, but it can also be quite funny at times.” that also occurs 5 times with an average annotation score of 1.4 which is incredibly high for this prompt. We agree that this lack of diversity hampers evaluation on lower temperatures, and note that our evaluations work best on diverse outputs.

\section{Full prompt list}
\label{app:prompt_list}
We show all the prompts we designed in Table~\ref{tab:prompts}. Our prompts used for temperature and model sensitivity experiments and other LLM experiments are in Table~\ref{tab:selected_prompts}
\clearpage
\onecolumn
\begin{longtable}{p{.15\textwidth} p{.15\textwidth} p{.63\textwidth}} 
\toprule
\textbf{Aspect} & \textbf{Variation} & \textbf{Prompt} \\
\midrule \midrule
\endhead

\multirow{24}{*}{\textbf{Writing Style}} & \multirow{6}{*}{\textbf{Base}} & Write a functional passage about sunsets:\\
& & Write a flowery passage about sunsets:\\
& & Write a functional passage about strawberries:\\
& & Write a flowery passage about strawberries:\\
& & Write a functional passage about writing a paper:\\
& & Write a flowery passage about writing a paper:\\
\cmidrule(lr){2-3}
& \multirow{6}{*}{\textbf{Template 2}} & Write a passage with a functional writing style about sunsets:\\
& & Write a passage with a flowery writing style about sunsets:\\
& & Write a passage with a functional writing style about strawberries:\\
& & Write a passage with a flowery writing style about strawberries:\\
& & Write a passage with a functional writing style about writing a paper:\\
& & Write a passage with a flowery writing style about writing a paper:\\
\cmidrule(lr){2-3}
& \multirow{12}{*}{\textbf{Advanced}} & Write a candid passage about sunsets:\\
& & Write a prosaic passage about sunsets:\\
& & Write an ornate passage about sunsets:\\
& & Write a poetic passage about sunsets:\\
& & Write a candid passage about strawberries:\\
& & Write a prosaic passage about strawberries:\\
& & Write an ornate passage about strawberries:\\
& & Write a poetic passage about strawberries:\\
& & Write a candid passage about writing a paper:\\
& & Write a prosaic passage about writing a paper:\\
& & Write an ornate passage about writing a paper:\\
& & Write a poetic passage about writing a paper:\\
\midrule
\multirow{32}{*}{\textbf{Tone}} & \multirow{4}{*}{\textbf{Base}} & Write a dramatic passage about love:\\
& & Write a humorous passage about love:\\
& & Write an optimistic passage about love:\\
& & Write a sad passage about love:\\
\cmidrule(lr){2-3}
& \multirow{4}{*}{\textbf{Subject 2}} & Write a dramatic passage about life:\\
& & Write a humorous passage about life:\\
& & Write a optimistic passage about life:\\
& & Write a sad passage about life:\\
\cmidrule(lr){2-3}
& \multirow{4}{*}{\textbf{Subject 3}} & Write a dramatic passage about humanity:\\
& & Write a humorous passage about humanity:\\
& & Write a optimistic passage about humanity:\\
& & Write a sad passage about humanity:\\
\cmidrule(lr){2-3}
& \multirow{4}{*}{\textbf{Template 2}} & Write a passage about love with a dramatic tone:\\
& & Write a passage about love with a humorous tone:\\
& & Write a passage about love with an optimistic tone:\\
& & Write a passage about love with a sad tone:\\
\cmidrule(lr){2-3}
& \multirow{4}{*}{\textbf{Template 3}} & Create a dramatic passage about love:\\
& & Create a humorous passage about love:\\
& & Create an optimistic passage about love:\\
& & Create a sad passage about love:\\
\cmidrule(lr){2-3}
\pagebreak
& \multirow{8}{*}{\textbf{Advanced}} & Write an emotional passage about love:\\
& & Write a nostalgic passage about love:\\
& & Write an ironic passage about love:\\
& & Write a satirical passage about love:\\
& & Write an uplifting passage about love:\\
& & Write an inspirational passage about love:\\
& & Write a bleak passage about love:\\
& & Write a grim passage about love:\\
\cmidrule(lr){2-3}
& \multirow{4}{*}{\textbf{Useful}} & Write a formal passage about love:\\
& & Write an informal passage about love:\\
& & Write an assertive passage about love:\\
& & Write a passive-aggressive passage about love:\\
\midrule
\multirow{27}{*}{\textbf{Mood}} & \multirow{4}{*}{\textbf{Base}} & Write a passage about love that makes the reader feel angry:\\
& & Write a passage about love that makes the reader feel fearful:\\
& & Write a passage about love that makes the reader feel happy:\\
& & Write a passage about love that makes the reader feel sad:\\
\cmidrule(lr){2-3}
& \multirow{4}{*}{\textbf{Subject 2}} & Write a passage about life that makes the reader feel angry:\\
& & Write a passage about life that makes the reader feel fearful:\\
& & Write a passage about life that makes the reader feel happy:\\
& & Write a passage about life that makes the reader feel sad:\\
\cmidrule(lr){2-3}
& \multirow{4}{*}{\textbf{Subject 3}} & Write a passage about humanity that makes the reader feel angry:\\
& & Write a passage about humanity that makes the reader feel fearful:\\
& & Write a passage about humanity that makes the reader feel happy:\\
& & Write a passage about humanity that makes the reader feel sad:\\
\cmidrule(lr){2-3}
& \multirow{4}{*}{\textbf{Template 2}} & Write a passage about love with an angry mood:\\
& & Write a passage about love with a fearful mood:\\
& & Write a passage about love with a happy mood:\\
& & Write a passage about love with a sad mood:\\
\cmidrule(lr){2-3}
& \multirow{4}{*}{\textbf{Template 3}} & Create a passage about love that makes the reader feel angry:\\
& & Create a passage about love that makes the reader feel fearful:\\
& & Create a passage about love that makes the reader feel happy:\\
& & Create a passage about love that makes the reader feel sad:\\
\cmidrule(lr){2-3}
& \multirow{4}{*}{\textbf{Template 4}} &  Write a passage about love that makes the reader feel anger:\\
& & Write a passage about love that makes the reader feel fear:\\
& & Write a passage about love that makes the reader feel happiness:\\
& & Write a passage about love that makes the reader feel sadness:\\
\cmidrule(lr){2-3}
& \multirow{7}{*}{\textbf{Advanced}} & Write a passage about love that makes the reader feel envious:\\
& & Write a passage about love that makes the reader feel anxious:\\
& & Write a passage about love that makes the reader feel proud:\\
& & Write a passage about love that makes the reader feel regretful:\\
& & Write a passage about love that makes the reader feel surprised:\\
& & Write a passage about love that makes the reader feel loved:\\
& & Write a passage about love that makes the reader feel disgusted:\\
\midrule
\pagebreak
\multirow{10}{*}{\textbf{Characterization}} & \multirow{2}{*}{\textbf{Base}} & Write a story about lovers with indirect characterization:\\
& & Write a story about lovers with direct characterization:\\
\cmidrule(lr){2-3}
& \multirow{2}{*}{\textbf{Subject 2}}  & Write a story about cats with indirect characterization:\\
& & Write a story about cats with direct characterization:\\
\cmidrule(lr){2-3}
& \multirow{2}{*}{\textbf{Subject 3}}  & Write a story about survivors with indirect characterization:\\
& & Write a story about survivors with direct characterization:\\
\cmidrule(lr){2-3}
& \multirow{2}{*}{\textbf{Template 2}}  & Write a story about lovers where the characters are described directly:\\
& & Write a story about lovers where the characters are described indirectly:\\
\cmidrule(lr){2-3}
& \multirow{2}{*}{\textbf{Template 3}}  & Create a story about lovers with indirect characterization:\\
& & Create a story about lovers with direct characterization:\\
\midrule
\multirow{10}{*}{\textbf{Pacing}} & \multirow{2}{*}{\textbf{Base}} & Write a fast-paced story about lovers:\\
& & Write a slow-paced story about lovers:\\
\cmidrule(lr){2-3}
& \multirow{2}{*}{\textbf{Subject 2}} & Write a fast-paced story about cats:\\
& & Write a slow-paced story about cats:\\
\cmidrule(lr){2-3}
& \multirow{2}{*}{\textbf{Subject 3}} & Write a fast-paced story about survivors:\\
& & Write a slow-paced story about survivors:\\
\cmidrule(lr){2-3}
& \multirow{2}{*}{\textbf{Template 2}} & Write a story about lovers that is fast-paced:\\
& & Write a story about lovers that is slow-paced:\\
\cmidrule(lr){2-3}
& \multirow{2}{*}{\textbf{Template 3}} & Create a fast-paced story about lovers:\\
& & Create a slow-paced story about lovers:\\
\midrule
\multirow{35}{*}{\textbf{Plot}} & \multirow{7}{*}{\textbf{Base}} & Write a story about lovers with an "Overcoming the Monster" plot:\\
& & Write a story about lovers with a "Rags to Riches" plot:\\
& & Write a story about lovers with a "The Quest" plot:\\
& & Write a story about lovers with a "Voyage and Return" plot:\\
& & Write a story about lovers with a "Comedy" plot:\\
& & Write a story about lovers with a "Tragedy" plot:\\
& & Write a story about lovers with a "Rebirth" plot:\\
\cmidrule(lr){2-3}
& \multirow{7}{*}{\textbf{Subject 2}} & Write a story about cats with an "Overcoming the Monster" plot:\\
& & Write a story about cats with a "Rags to Riches" plot:\\
& & Write a story about cats with a "The Quest" plot:\\
& & Write a story about cats with a "Voyage and Return" plot:\\
& & Write a story about cats with a "Comedy" plot:\\
& & Write a story about cats with a "Tragedy" plot:\\
& & Write a story about cats with a "Rebirth" plot:\\
\cmidrule(lr){2-3}
& \multirow{7}{*}{\textbf{Subject 3}} & Write a story about survivors with an "Overcoming the Monster" plot:\\
& & Write a story about survivors with a "Rags to Riches" plot:\\
& & Write a story about survivors with a "The Quest" plot:\\
& & Write a story about survivors with a "Voyage and Return" plot:\\
& & Write a story about survivors with a "Comedy" plot:\\
& & Write a story about survivors with a "Tragedy" plot:\\
& & Write a story about survivors with a "Rebirth" plot:\\
\cmidrule(lr){2-3}
\pagebreak
& \multirow{7}{*}{\textbf{Subject 4}} & Write a story with an "Overcoming the Monster" plot:\\
& & Write a story with a "Rags to Riches" plot:\\
& & Write a story with a "The Quest" plot:\\
& & Write a story with a "Voyage and Return" plot:\\
& & Write a story with a "Comedy" plot:\\
& & Write a story with a "Tragedy" plot:\\
& & Write a story with a "Rebirth" plot:\\
\cmidrule(lr){2-3}
& \multirow{7}{*}{\textbf{Template 2}} & Create a story about lovers with an "Overcoming the Monster" plot:\\
& & Create a story about lovers with a "Rags to Riches" plot:\\
& & Create a story about lovers with a "The Quest" plot:\\
& & Create a story about lovers with a "Voyage and Return" plot:\\
& & Create a story about lovers with a "Comedy" plot:\\
& & Create a story about lovers with a "Tragedy" plot:\\
& & Create a story about lovers with a "Rebirth" plot:\\
\midrule
\multirow{36}{*}{\textbf{Genre}} & \multirow{6}{*}{\textbf{Base}} & Write a historical fiction story about lovers:\\
& & Write a literary fiction story about lovers:\\
& & Write a mystery story about lovers:\\
& & Write a science fiction story about lovers:\\
& & Write a dystopian story about lovers:\\
& & Write a horror story about lovers:\\
\cmidrule(lr){2-3}
& \multirow{6}{*}{\textbf{Subject 2}} & Write a historical fiction story about cats:\\
& & Write a literary fiction story about cats:\\
& & Write a mystery story about cats:\\
& & Write a science fiction story about cats:\\
& & Write a dystopian story about cats:\\
& & Write a horror story about cats:\\
\cmidrule(lr){2-3}
& \multirow{6}{*}{\textbf{Subject 3}} & Write a historical fiction story about survivors:\\
& & Write a literary fiction story about survivors:\\
& & Write a mystery story about survivors:\\
& & Write a science fiction story about survivors:\\
& & Write a dystopian story about survivors:\\
& & Write a horror story about survivors:\\
\cmidrule(lr){2-3}
& \multirow{6}{*}{\textbf{Subject 4}} & Write a historical fiction story:\\
& & Write a literary fiction story:\\
& & Write a mystery story:\\
& & Write a science fiction story:\\
& & Write a dystopian story:\\
& & Write a horror story:\\
\cmidrule(lr){2-3}
& \multirow{6}{*}{\textbf{Template 2}} & Write a story about lovers in a historical fiction genre:\\
& & Write a story about lovers in a literary fiction genre:\\
& & Write a story about lovers in a mystery genre:\\
& & Write a story about lovers in a science fiction genre:\\
& & Write a story about lovers in a dystopian genre:\\
& & Write a story about lovers in a horror genre:\\
\cmidrule(lr){2-3}
& \multirow{6}{*}{\textbf{Template 3}} & Create a historical fiction story about lovers:\\
& & Create a literary fiction story about lovers:\\
& & Create a mystery story about lovers:\\
& & Create a science fiction story about lovers:\\
& & Create a dystopian story about lovers:\\
& & Create a horror story about lovers:\\
\midrule
\multirow[c]{36}{*}{\textbf{Numerical}} & \multirow{27}{*}{\textbf{Base}}  & Write a sentence with five words about love:\\
& & Write a sentence with five words about cats:\\
& & Write a sentence with five words about running:\\
& & Write a sentence with ten words about love:\\
& & Write a sentence with ten words about cats:\\
& & Write a sentence with ten words about running:\\
& & Write a sentence with twenty words about love:\\
& & Write a sentence with twenty words about cats:\\
& & Write a sentence with twenty words about running:\\
& & Write a paragraph with five sentences about love:\\
& & Write a paragraph with five sentences about cats:\\
& & Write a paragraph with five sentences about running:\\
& & Write a paragraph with ten sentences about love:\\
& & Write a paragraph with ten sentences about cats:\\
& & Write a paragraph with ten sentences about running:\\
& & Write a paragraph with twenty sentences about love:\\
& & Write a paragraph with twenty sentences about cats:\\
& & Write a paragraph with twenty sentences about running:\\
& & Write a passage with five paragraphs about love:\\
& & Write a passage with five paragraphs about cats:\\
& & Write a passage with five paragraphs about running:\\
& & Write a passage with ten paragraphs about love:\\
& & Write a passage with ten paragraphs about cats:\\
& & Write a passage with ten paragraphs about running:\\
& & Write a passage with twenty paragraphs about love:\\
& & Write a passage with twenty paragraphs about cats:\\
& & Write a passage with twenty paragraphs about running:\\
\cmidrule(lr){2-3}
& \multirow{27}{*}{\textbf{Template 2}} & Write a sentence about love with 5 words:\\
& & Write a sentence about cats with 5 words:\\
& & Write a sentence about running with 5 words:\\
& & Write a sentence about love with 10 words:\\
& & Write a sentence about cats  with 10 words:\\
& & Write a sentence about running with 10 words:\\
& & Write a sentence about love with 20 words:\\
& & Write a sentence about cats with 20 words:\\
& & Write a sentence about running with 20 words:\\
& & Write a paragraph about love with 5 sentences:\\
& & Write a paragraph about cats with 5 sentences:\\
& & Write a paragraph about running with 5 sentences:\\
& & Write a paragraph about love  with 10 sentences:\\
& & Write a paragraph about cats with 10 sentences:\\
& & Write a paragraph about running with 10 sentences:\\
& & Write a paragraph about love with 20 sentences:\\
& & Write a paragraph about cats with 20 sentences:\\
& & Write a paragraph about running with 20 sentences:\\
& & Write a passage about love with 5 paragraphs:\\
& & Write a passage about cats with 5 paragraphs:\\
& & Write a passage about running with 5 paragraphs:\\
& & Write a passage about love with 10 paragraphs:\\
& & Write a passage about cats with 10 paragraphs:\\
& & Write a passage about running with 10 paragraphs:\\
& & Write a passage about love with 20 paragraphs:\\
& & Write a passage about cats  with 20 paragraphs:\\
& & Write a passage about running  with 20 paragraphs:\\
\midrule
\multirow{36}{*}{\textbf{Descriptive}} & \multirow{18}{*}{\textbf{Base}}  & Write a short passage about love:\\
& & Write a brief passage about love:\\
& & Write a concise passage about love:\\
& & Write a long passage about love:\\
& & Write a detailed passage about love:\\
& & Write an in-depth passage about love:\\
& & Write a short passage about cats:\\
& & Write a brief passage about cats:\\
& & Write a concise passage about cats:\\
& & Write a long passage about cats:\\
& & Write a detailed passage about cats:\\
& & Write an in-depth passage about cats:\\
& & Write a short passage about running:\\
& & Write a brief passage about running:\\
& & Write a concise passage about running:\\
& & Write a long passage about running:\\
& & Write a detailed passage about running:\\
& & Write an in-depth passage about running:\\
\cmidrule(lr){2-3}
& \multirow{18}{*}{\textbf{Template 2}} & Write a passage about love that is short:\\
& & Write a passage about love that is brief:\\
& & Write a passage about love that is concise:\\
& & Write a passage about love that is long:\\
& & Write a passage about love that is detailed:\\
& & Write a passage about love that is in-depth:\\
& & Write a passage about cats that is short:\\
& & Write a passage about cats that is brief:\\
& & Write a passage about cats that is concise:\\
& & Write a passage about cats that is long:\\
& & Write a passage about cats that is detailed:\\
& & Write a passage about cats that is in-depth:\\
& & Write a passage about running that is short:\\
& & Write a passage about running that is brief:\\
& & Write a passage about running that is concise:\\
& & Write a passage about running that is long:\\
& & Write a passage about running that is detailed:\\
& & Write a passage about running that is in-depth:\\
\midrule
\pagebreak
\multirow{36}{*}{\textbf{Functional}} & \multirow{8}{*}{\textbf{Code}}  & Code Python code that plays the game of war:\\
& & Code Python code that sums two integers up:\\
& & Code C code that plays the game of war:\\
& & Code C code that sums two integers up:\\
& & Write Python code that plays the game of war:\\
& & Write Python code that sums two integers up:\\
& & Write C code that plays the game of war:\\
& & Write C code that sums two integers up:\\
\cmidrule(lr){2-3}
& \multirow{6}{*}{\textbf{Email}} & Write an email to my teacher:\\
& & Write an email to my teacher asking for help on math homework:\\
& & Write an email to my boyfriend:\\
& & Write an email to my boyfriend to arrange a date this Saturday:\\
& & Write an email to my client:\\
& & Write an email to my client requesting a copy of the updated contract:\\
\cmidrule(lr){2-3}
& \multirow{10}{*}{\textbf{Academic Paper}} & Write a properly formatted academic paper on artificial intelligence:\\
& & Write an academic paper on artificial intelligence in the proper format:\\
& & Write a properly formatted academic paper on the flaws of GPT-3:\\
& & Write an academic paper on the flaws of GPT-3 in the proper format:\\
& & Write a properly formatted academic paper on strategies our society can adopt to recover after the global pandemic as quickly and painlessly as possible:\\
& & Write an academic paper on strategies our society can adopt to recover after the global pandemic as quickly and painlessly as possible in the proper format:\\
\bottomrule
\caption{The full list of the prompts.}
\label{tab:prompts}
\end{longtable}

\begin{table*}[h]
\scalebox{1}{
    \begin{tabular}{@{}cl@{}}
    \toprule
    
    0 & Write a flowery passage about sunsets:\\
    1 & Write a humorous passage about love:\\
    2 & Write a passage about love that makes the reader feel fearful:\\
    3 & Write a story about lovers with indirect characterization:\\
    4 & Write a fast-paced story about lovers:\\
    5 & Write a story about lovers with a "Tragedy" plot:\\
    6 & Write a historical fiction story about lovers:\\
    \bottomrule
    \end{tabular}
}
\caption{\textbf{Selected prompts for additional experiments}}
\label{tab:selected_prompts}
\end{table*}

\clearpage
\twocolumn

\subsection{Example Mitigations}
\label{app:example mitigations}

\noindent \textbf{Example Definition:}

A humorous tone is a light, playful, and funny tone.

Write a humorous passage about love: \\

\noindent \textbf{Example Demonstration:}

Write a humorous passage about life:

If life gives you lemons, make lemonade, sell it in a rich neighborhood, invest all the money in crypto, and retire before you’re 30. At least that's what I heard on TikTok. Come to think of it, if I put just a little effort into my lemonade stand 10 years ago,  I would be traveling the world right now instead of writing jokes for a living.

Write a humorous passage about love: \\

\noindent \textbf{Example Explanation:}

Write a humorous passage about life:

If life gives you lemons, make lemonade, sell it in a rich neighborhood, invest all the money in crypto, and retire before you’re 30. At least that's what I heard on TikTok. Come to think of it, if I put just a little effort into my lemonade stand 10 years ago,  I would be traveling the world right now instead of writing jokes for a living.

Explanation: This passage is humorous because it takes a common proverb and adds a crazy and unrealistic twist. It also uses a deadpan tone for a completely unrealistic scenario, which is funny due to the disparity between tone and subject matter.

Write a humorous passage about love:

\end{document}